%% file: main.tex
\documentclass[10pt,journal,compsoc]{IEEEtran}
\usepackage[nocompress]{cite}
\usepackage[pdftex]{graphicx}
\graphicspath{{figures/}}
\DeclareGraphicsExtensions{.pdf,.jpeg,.png}
\usepackage{amsmath,bm}
\usepackage{array}
\usepackage{url}
\usepackage[x11names]{xcolor}
\usepackage{tabulary,xfrac,enumitem}
\usepackage{booktabs}       % professional-quality tables
\usepackage{amsfonts}       % blackboard math symbols
\usepackage{nicefrac}       % compact symbols for 1/2, etc.
\usepackage{microtype}      % microtypography
\usepackage{amssymb}
\usepackage{amsmath}
\usepackage{caption, subcaption}
\usepackage{tabularx}
\usepackage{multicol}
\usepackage{multirow} 
\usepackage{textcomp}
\usepackage{makecell}
\usepackage{colortbl}
\usepackage{pifont}% http://ctan.org/pkg/pifont
\usepackage{xspace}
\definecolor{infraLinkColor}{RGB}{150,0,20}
\definecolor{infraCiteColor}{RGB}{0,110,60}
\definecolor{infraUrlColor}{RGB}{0,70,120}
\usepackage[
  pagebackref=true,
  breaklinks=true,
  colorlinks=true,
  linkcolor=infraLinkColor,
  citecolor=infraCiteColor,
  urlcolor=infraUrlColor,
  bookmarks=false
]{hyperref}

\usepackage{xcolor}
\usepackage{color, colortbl}
\usepackage{verbatim}
\usepackage{marvosym}
\usepackage{ragged2e}

\usepackage[capitalize]{cleveref}
\crefname{section}{Sec.}{Secs.}
\Crefname{section}{Section}{Sections}

\Crefname{table}{Table}{Tables}
\crefname{table}{Tab.}{Tabs.}
\Crefname{figure}{Fig.}{Figs.}
\crefname{figure}{Fig.}{Figs.}

\renewcommand{\figurename}{Figure}

\newcommand{\nameplus}{\mbox{ProSD-Occ}}
\newcommand{\rev}[1]{#1}

\newcommand{\cmark}{\ding{51}\xspace}

\newcommand{\xmark}{\ding{55}\xspace}

\makeatletter
\newcommand{\thickhline}{%
    \noalign {\ifnum 0=`}\fi \hrule height 0.5pt
    \futurelet \reserved@a \@xhline
}
\def\section{\@startsection{section}{1}{\z@}{-3.5ex plus -2ex minus -1.5ex}%
{0.7ex plus 1ex minus 0ex}{\normalfont\sublargesize\sffamily\bfseries}}%
\makeatother

\begin{document}
\title{InfraOcc: An Infrastructure Occupancy Benchmark with Static-to-Dynamic Reasoning}

\author{Lei~Yang\textsuperscript{$\dagger$}, Xiaokai~Bai\textsuperscript{$\dagger$}, Boqi~Li, Chunmian Lin, Li~Wang, Ziying Song, Jiahuan Zhang, Enhui Ma \\ Haibao Yu, Jiaqi Ma, Kaicheng~Yu
\IEEEcompsocitemizethanks{
\IEEEcompsocthanksitem \textsuperscript{$\dagger$} equal contribution.
\IEEEcompsocthanksitem Lei Yang is with the School of Mechanical and
Aerospace Engineering, Nanyang Technological University, Singapore. Email: lei.yang@ntu.edu.sg.
\IEEEcompsocthanksitem Xiaokai Bai is with the College of Information Science
and Electronic Engineering, Zhejiang University, Hangzhou 310027, China.
\IEEEcompsocthanksitem Boqi Li and Chunmian Lin are with the School of
Transportation Science and Engineering, Beihang University, Beijing, China.
\IEEEcompsocthanksitem Li Wang is with the School of Mechanical Engineering,
Beijing Institute of Technology, Beijing, China.
\IEEEcompsocthanksitem Ziying Song is with the School of Artificial
Intelligence, Yanshan University, China.
\IEEEcompsocthanksitem Jiahuan Zhang, Enhui Ma, and Kaicheng Yu are with
Westlake University, Hangzhou, China.
\IEEEcompsocthanksitem Haibao Yu is with the University of Hong Kong,
Hong Kong, China.
\IEEEcompsocthanksitem Jiaqi Ma is with the University of California,
Los Angeles, CA, USA.
\IEEEcompsocthanksitem Corresponding authors: Boqi Li (boqili@buaa.edu.cn), Li Wang (wangli\_bit@bit.edu.cn), and Kaicheng Yu (kyu@westlake.edu.cn). 
kyu@westlake.edu.cn
}
}

\markboth{IEEE Transactions on Pattern Analysis and Machine Intelligence}%
{Yang \MakeLowercase{\textit{et al.}}: InfraOcc: An Infrastructure Occupancy Benchmark with Static-to-Dynamic Reasoning}

\IEEEtitleabstractindextext{%
\begin{abstract}
\justifying
Fixed-viewpoint infrastructure sensors repeatedly observe the same traffic
space, making roadside 3D occupancy structurally different from ego-vehicle
perception: a near-persistent static scaffold is overlaid with sparse,
short-lived dynamic events. Existing occupancy benchmarks and methods, however,
are built around moving ego vehicles and neither measure nor exploit this
structure, instead treating occupancy as flat one-shot voxel classification. We
address this gap from both data and model perspectives. We build
\textbf{InfraOcc}, to our knowledge, the first real-world infrastructure-side semantic occupancy
benchmark, with dense voxel annotations for 290 multi-modal sequences in a fixed
roadside frame, a static-dynamic decoupled annotation pipeline, unified
\rev{camera-only, LiDAR-only, and multi-modal} evaluation, and diagnostics for static and
dynamic occupancy. InfraOcc shows that static infrastructure fills 97.3\% of
occupied voxels and persists across frames, whereas dynamic participants have a
\rev{median occupied-frame ratio} of only 1.8\% per location, revealing a structural
static-dynamic asymmetry beyond semantic long-tailedness. We further propose
\nameplus{}, which reformulates occupancy as progressive static-to-dynamic
evidence reasoning: it explains persistent layout, exposes residual dynamic
evidence under static-confidence guidance, and recomposes static, dynamic, and
free-space evidence into a unified field. \rev{\nameplus{} ranks first in overall, dynamic, static, and geometric
occupancy on every track, e.g.,
a 23.5\% relative camera-only dynamic-mIoU gain over the strongest baseline
and 65.87 multi-modal overall mIoU,} establishing fixed-viewpoint
roadside occupancy as a distinct problem with its own reasoning paradigm.
\rev{The benchmark and code will be publicly available at
\url{https://github.com/yanglei18/InfraOcc}.}
\end{abstract}

% Note that keywords are not normally used for peer review papers.
}

\maketitle

\IEEEdisplaynontitleabstractindextext

\IEEEpeerreviewmaketitle

\IEEEraisesectionheading{\section{Introduction}\label{sec:introduction}}
\input{latex/1-intro}

\section{Related Work}\label{sec:related_work}
\input{latex/2-relatedworks}

\section{InfraOcc Benchmark}
\label{sec:infraocc_benchmark}
\input{latex/3-benchmark}

\input{latex/table_macros}

\section{ProSD-Occ Framework}
\label{sec:method}
\input{latex/3-method}

\section{Experiments}
\label{sec:exps}
\input{latex/4-exps}

\section{Conclusion} 
\label{sec:conclusion}
\input{latex/5-conclusion}

\section*{Acknowledgments}
This work was supported by the Zhejiang Leading Innovative and Entrepreneur Team Introduction Program (2024R01007), the National Natural Science Foundation of China (52502496, 62403389), the Beijing Natural Science Foundation (L2609087), the Natural Science Foundation of Zhejiang, China (QKWL25F0301), and the Natural Science Foundation of Chongqing, China (CSTB2025NSCQ-GPX0413).

\ifCLASSOPTIONcaptionsoff
  \newpage
\fi

% \clearpage
\bibliographystyle{IEEEtran}
\bibliography{IEEEabrv, egbib}

\end{document}

%% file: latex/1-intro.tex
%%%%%%%%% BODY TEXT

\input{latex/fig/dataset-sample.tex}

\IEEEPARstart{I}{nfrastructure}-side perception is becoming an important pillar
of autonomous driving and intelligent transportation: cameras and LiDARs fixed
at intersections and road segments continuously observe long-term traffic
situations beyond the field of view of on-board sensors, providing scene-level
3D understanding for vehicles, traffic management, and safety monitoring. In
recent years, semantic occupancy prediction, which jointly estimates geometric
occupancy and semantic categories in a regular voxel space and unifies static
layout, dynamic participants, and free space within a single 3D
representation, has become an important paradigm for driving scene
understanding~\cite{OpenOccupancy,Occ3D,TPVFormer,SurroundOcc}. However, these
representations and benchmarks are almost exclusively designed for vehicle-side
ego-centric perception, where the occupancy volume is defined around a moving
ego vehicle and the viewpoint and coordinate reference change over
time~\cite{nuScenes,Occ3D,OpenOccupancy,SurroundOcc}. In contrast, roadside
cameras and LiDARs are fixed in place and repeatedly observe the same traffic
space over long periods~\cite{Rope3D,DAIRV2X,V2XReal,V2XPnP,RCooper,TUMTrafV2X}.
\rev{This fixed viewpoint is not a mere change of camera placement: as we
quantify in Sec.~\ref{sec:benchmark_statistics}, it fundamentally alters the
problem structure of 3D occupancy prediction. On
the vehicle side, occupancy prediction estimates a continuously changing
instantaneous scene frame by frame. On the roadside, by contrast, the same
space is observed repeatedly, and the occupancy field reduces to a
near-persistent static spatial
scaffold overlaid with sparse, short-lived dynamic traffic events.}
Fixed-viewpoint roadside occupancy should therefore be treated as a distinct
problem with its own structure, rather than a viewpoint transfer of
vehicle-side occupancy.

Existing research, however, advances along two main lines, neither of which
addresses this structure. \rev{First, mainstream semantic occupancy
benchmarks and methods are ego-centric. Existing methods build BEV/voxel
features through view
transformation and depth lifting~\cite{BEVDet,BEVDepth}, aggregate multi-view
and temporal context with transformer queries~\cite{BEVFormer,TPVFormer}, or
improve voxel reasoning and sparse/compact
representations~\cite{VoxFormer,OccFormer,SurroundOcc,SparseOcc,GaussianFormer},
yet their voxel space is always defined around the ego vehicle with a moving
coordinate frame. \rev{As a result, they} neither characterize the long-term spatial
structure under a fixed roadside viewpoint nor provide data and diagnostic
metrics that measure static-dynamic organization.} Second, existing infrastructure-side and
cooperative perception studies are largely object-centric: roadside and
cooperative perception
datasets~\cite{Rope3D,DAIRV2X,V2XReal,V2XPnP,RCooper,TUMTrafV2X,OPV2V,V2XRadar} and
methods targeting roadside-camera geometry modeling, ground priors, scenario
generalization, calibration robustness, and cooperative feature
fusion~\cite{BEVHeight,BEVHeightPP,MonoGAE,SGV3D,SIFormer,V2XViT,Where2comm}
mostly focus on 3D detection, tracking, or cooperative detection, yielding
sparse object boxes rather than a dense scene description that jointly
represents static infrastructure, dynamic participants, and free space. A few
cooperative semantic occupancy works have begun to connect V2X perception with
dense scene completion~\cite{CollaborativeSemanticOcc,SyntheticCollaborativeOcc},
but they mainly target connected-vehicle cooperation or synthetic settings, and
dense semantic occupancy under real fixed roadside sensors remains unexplored.
As a result, the core structure of ``persistent static versus transient
dynamic'' under a fixed viewpoint has neither a benchmark to measure it nor a
method to exploit it.

\rev{To fill this gap, we build \textbf{InfraOcc}
(Fig.~\ref{fig:dataset_sample}), which is, to our knowledge, the first
semantic occupancy benchmark constructed from real fixed roadside sensors.} Constructing roadside
occupancy annotations is not a simple conversion of existing 3D annotations into
voxel labels: fixed roadside LiDAR is limited by a single viewpoint and
occlusions, making it hard to fully cover persistent static background such as
roads, buildings, vegetation, and \rev{barriers}, while dynamic objects in a single
roadside scan are often sparse and \rev{incomplete at object boundaries}. InfraOcc therefore
adopts a static-dynamic decoupled annotation pipeline: a dynamic branch
aggregates dynamic geometry from roadside LiDAR sequences and temporally
consistent object tracklets, a static branch completes the static background
using the moving viewpoints of vehicle-side LiDAR, and the two are then
recomposed in a fixed roadside coordinate system with visibility-aware labeling.
We emphasize that vehicle-side LiDAR is used only for annotation construction;
benchmark inputs are always restricted to roadside cameras, roadside LiDAR, or
their combination. InfraOcc further provides a unified camera-only, LiDAR-only,
and multi-modal evaluation protocol, together with diagnostic metrics that
separate static and dynamic occupancy.

\rev{Statistical analysis of InfraOcc provides a quantitative
characterization of the static-dynamic structure under a fixed viewpoint.} \rev{Temporally, measured by
the occupied-frame ratio (the fraction of frames in which a location is
occupied), static infrastructure is highly persistent (a median of
100\%), whereas dynamic participants reach a median of only 1.8\% per
location and a 95th percentile of about
16.5\%}; in terms of semantic frequency, static infrastructure accounts for
97.3\% of occupied voxels while dynamic participants account for only 2.7\%.
\rev{This shows that the fundamental difficulty of roadside occupancy is not
merely semantic long-tailedness but a structural static-dynamic asymmetry.}
Under this structure, flat one-shot
voxel classification biases the model toward frequent and spatially dominant
static patterns, weakening sparse dynamic cues during voxel feature construction
and 3D context aggregation; simple loss
reweighting~\cite{FocalLoss,ClassBalancedLoss} can alleviate class imbalance but
cannot change the way static and dynamic evidence are homogeneously organized.
\input{latex/tab/dataset-comparison.tex}

Building on this, we propose \textbf{ProSD-Occ}, a progressive
static-to-dynamic reasoning framework that reformulates fixed-viewpoint
occupancy prediction from ``one-shot classification of all voxels'' into
``step-by-step reasoning ordered by evidence reliability.'' \rev{Its core idea
directly mirrors the structural asymmetry above: the model should first
explain the stable, reusable part of the scene and then infer what this
explanation cannot account for, rather than
letting sparse dynamic cues be overwhelmed by spatially dominant static patterns
in a single classification.} \rev{Concretely, ProSD-Occ first establishes a
sample-adaptive explanation of the persistent static layout (Static Layout
Reasoner), then exposes residual evidence that the layout cannot sufficiently
account for---often sparse yet safety-critical dynamic objects (Static-guided
Residual Modulation and Residual Dynamic Predictor)---and finally recomposes
static, dynamic, and free-space evidence into a unified semantic occupancy
field (Semantic Recomposition).} \rev{In this way,
ProSD-Occ turns the static-dynamic asymmetry of the fixed viewpoint from a
source of bias into an organizing principle: it
retains the contextual value of stable static layout while preventing it from
dominating the prediction of sparse dynamic objects.}

\rev{On InfraOcc, ProSD-Occ ranks first in overall, dynamic, static, and
geometric occupancy on every track, e.g., a 23.5\% relative dynamic-mIoU
gain over the strongest camera baseline (28.97 vs. 23.45) and 65.87 overall
mIoU in the multi-modal track. The margins persist under coordinate
re-anchoring, indicating that static-to-dynamic evidence organization
provides consistent, evidence-driven benefits for fixed-view occupancy
prediction.}
The main contributions of this paper are summarized as follows:
\begin{itemize}
\item \rev{We establish \textbf{InfraOcc}, to our knowledge the first
real-world benchmark for fixed-viewpoint semantic occupancy: a
static-dynamic decoupled annotation pipeline turns object-level roadside
data into dense voxel labels in a fixed roadside frame, with unified
camera-only, LiDAR-only, and multi-modal protocols and static/dynamic
diagnostics.}

\item \rev{We identify and quantify a defining property of fixed-viewpoint
occupancy: a \textbf{structural static-dynamic asymmetry} (97.3\% vs.
2.7\% of occupied voxels; near-permanent presence vs. a 1.8\% median
occupied-frame ratio). This difficulty goes beyond semantic long-tailedness
and is neither measured nor exploited by existing benchmarks and methods.}

\item \rev{We propose \textbf{ProSD-Occ}, which reformulates fixed-viewpoint
occupancy from flat one-shot classification into progressive
static-to-dynamic evidence reasoning, and ranks first in overall, dynamic,
static, and geometric occupancy on every InfraOcc track, with up to a
23.5\% relative dynamic-mIoU gain.}
\end{itemize}

%% file: latex/fig/dataset-sample.tex
\begin{figure*}[!t]
  \centering
  \includegraphics[width=\textwidth]{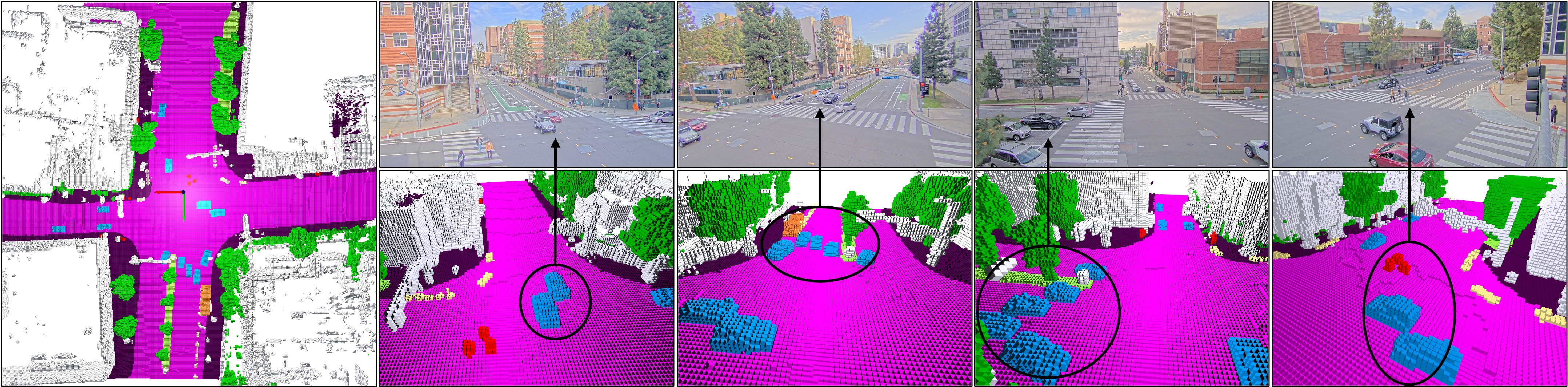}
  \caption{\rev{\textbf{A representative keyframe of the InfraOcc benchmark.}
  Left: dense voxel-level semantic occupancy annotation of the whole scene in
  the fixed roadside coordinate system (bird's-eye view). Right: the four
  synchronized roadside camera views (top) and the corresponding local 3D
  occupancy regions (bottom), where circles and arrows associate dynamic
  traffic participants across the two modalities. Persistent static layout
  dominates the scene, whereas dynamic participants occupy only sparse,
  transient local regions---the structural asymmetry studied in this paper.}}
  \label{fig:dataset_sample}
\end{figure*}

%% file: latex/tab/dataset-comparison.tex
% Benchmark comparison table.
% Scope: fixed-viewpoint dense occupancy setting.
% \cmark / \xmark are defined in main.tex; \pmark is a local partial-support mark.
% Sensor = roadside/infrastructure-side suite producing benchmark inputs;
% ego rows show the ego-vehicle suite; cooperative datasets additionally include
% vehicle agents (see Viewpoint).
\providecommand{\pmark}{\textcolor{gray}{$\triangle$}\xspace}
\begin{table*}[t]
\caption{\textbf{Comparison of representative 3D perception benchmarks across
viewpoint, sensing, and occupancy capabilities.} Among them, InfraOcc is the
only real-world benchmark that provides dense semantic occupancy under a fixed
roadside viewpoint with unified C/L/C+L evaluation and separate static/dynamic
(S/D) evaluation.
\emph{Sensor} denotes the roadside/infrastructure-side suite producing benchmark
inputs (ego rows show the ego-vehicle suite); cooperative datasets additionally
include vehicle agents, as indicated by \emph{Viewpoint}. C/L/R denote camera,
LiDAR, and radar. \rev{Det/Track/Pred/Occ denote detection, tracking, prediction, and occupancy
tasks; \emph{Temporal} indicates temporally continuous sequences; \emph{Occ.
States} denotes occupied/free/unobserved voxel labels.}
\cmark/\xmark/\pmark denote full/no/partial support.}
\label{tab:dataset_comparison}
\centering
\footnotesize
\setlength{\tabcolsep}{4pt}
\renewcommand{\arraystretch}{1.15}
\begin{tabular*}{\textwidth}{@{\extracolsep{\fill}}llllccccc@{}}
\toprule
Benchmark & Viewpoint & Sensor & Tasks & Real & Temporal & Dense Occ. & Occ. States & S/D Eval. \\
\midrule
\multicolumn{9}{@{}l}{\textit{Ego-centric occupancy}} \\
Occ3D~\cite{Occ3D}            & Ego & 6C\,+\,1L & Occ & \cmark & \cmark & \cmark & \cmark & \xmark \\
OpenOccupancy~\cite{OpenOccupancy} & Ego & 6C\,+\,1L & Occ & \cmark & \cmark & \cmark & \pmark & \xmark \\
SurroundOcc~\cite{SurroundOcc} & Ego & 6C       & Occ & \cmark & \cmark & \cmark & \xmark & \xmark \\
\midrule
\multicolumn{9}{@{}l}{\textit{Roadside / cooperative detection}} \\
Rope3D~\cite{Rope3D}          & Roadside        & 1C            & Det     & \cmark & \xmark & \xmark & \xmark & \xmark \\
DAIR-V2X~\cite{DAIRV2X}       & Coop.\,(V+I)    & 1C\,+\,1L     & Det     & \cmark & \cmark & \xmark & \xmark & \xmark \\
V2X-Real~\cite{V2XReal}       & Coop.\,(V+I)    & 4C\,+\,2L     & Det       & \cmark & \xmark & \xmark & \xmark & \xmark \\
V2XPnP~\cite{V2XPnP}          & Coop.\,(V+I)    & 4C\,+\,2L     & Det/Pred & \cmark & \cmark & \xmark & \xmark & \xmark \\
RCooper~\cite{RCooper}        & Roadside        & $\le$2C\,+\,2L & Det/Track & \cmark & \cmark & \xmark & \xmark & \xmark \\
TUMTraf-V2X~\cite{TUMTrafV2X} & Coop.\,(V+I)    & 3C\,+\,2L     & Det/Track & \cmark & \cmark & \xmark & \xmark & \xmark \\
V2X-Radar~\cite{V2XRadar}     & Coop.\,(V+I)    & 3C\,+\,1L\,+\,1R & Det  & \cmark & \cmark & \xmark & \xmark & \xmark \\
\midrule
\multicolumn{9}{@{}l}{\textit{Cooperative occupancy (synthetic)}} \\
CoHFF~\cite{CollaborativeSemanticOcc}      & Coop.\,(V) & 4C\,+\,1L & Occ & \xmark & \cmark & \cmark & \xmark & \xmark \\
Co3SOP~\cite{SyntheticCollaborativeOcc}    & Coop.\,(V) & 4C\,+\,1L & Occ & \xmark & \cmark & \cmark & \pmark & \xmark \\
\midrule
\textbf{InfraOcc (Ours)} & \textbf{Roadside-fixed} & \textbf{4C\,+\,2L} & \textbf{Occ} & \cmark & \cmark & \cmark & \cmark & \cmark \\
\bottomrule
\end{tabular*}
\end{table*}

%% file: latex/2-relatedworks.tex
\subsection{Semantic Occupancy Prediction}
\label{sec:related_occupancy}

Semantic occupancy prediction jointly estimates geometry and semantics in a
regular voxel space, giving autonomous driving a dense 3D scene representation.
Early semantic scene completion completed voxel semantics from partial indoor or
LiDAR observations, e.g., SSCNet~\cite{SSCNet} and
SemanticKITTI~\cite{SemanticKITTI}. Occ3D~\cite{Occ3D} and
OpenOccupancy~\cite{OpenOccupancy} then extended occupancy prediction to
large-scale driving scenes and standardized dense evaluation. Mainstream methods
advance through camera-side view transformation and voxel
construction~\cite{BEVDet,BEVDepth,BEVFormer,TPVFormer,COTR,HiSOP},
explicit geometry and camera-LiDAR fusion~\cite{VoxelNet,PointPillars,
BEVFusion,OccFormer,SurroundOcc,REO}, and efficient representation or supervision
schemes~\cite{SparseOcc,GaussianFormer,OccWorld,UniOcc}. \rev{Among them,
STCOcc~\cite{STCOcc} exploits sparse spatio-temporal cues and is the
strongest one-shot camera method in our experiments.} These works improve feature
construction, efficiency, and supervision, but remain ego-centric: voxel spaces
move with the vehicle, and outputs are treated as homogeneous voxel-wise
classification. \rev{None of them characterizes} fixed-view long-term spatial
structure or \rev{distinguishes} static layout from dynamic evidence. \rev{This paper
instead studies occupancy in a fixed roadside coordinate system, where the
voxel space is anchored to the scene rather than to a moving vehicle, and
treats the resulting static-dynamic structure as an explicit modeling
target.}

\subsection{Infrastructure-side and Cooperative Perception}
\label{sec:related_infrastructure}

Infrastructure-side and cooperative perception extend single-vehicle coverage
with roadside sensors or multi-agent communication. Datasets such as
Rope3D~\cite{Rope3D}, DAIR-V2X~\cite{DAIRV2X}, V2X-Real~\cite{V2XReal},
V2XPnP~\cite{V2XPnP}, TUMTraf-V2X~\cite{TUMTrafV2X},
RCooper~\cite{RCooper}, OPV2V~\cite{OPV2V}, and
V2X-Radar~\cite{V2XRadar} support roadside perception, vehicle-infrastructure
cooperation, and multi-agent collaboration. Methods study roadside geometry and
robust detection through height modeling~\cite{BEVHeight,BEVHeightPP}, ground
priors~\cite{MonoGAE}, scenario generalization~\cite{SGV3D}, calibration-aware
multi-sensor perception~\cite{SIFormer}, and cooperative feature
fusion~\cite{V2XViT,Where2comm,CoDriving}. These studies show the value of fixed infrastructure and
cooperative sensing under long-range or occluded conditions, but remain
object-centric and produce sparse boxes rather than dense static/dynamic/free
space scene descriptions. Recent cooperative occupancy works connect V2X
perception with dense completion~\cite{CollaborativeSemanticOcc,
SyntheticCollaborativeOcc}, but focus on multi-agent fusion or synthetic V2X
settings. Dense semantic occupancy under real fixed roadside sensors therefore
remains unexplored. \rev{InfraOcc fills this gap by upgrading real roadside sensing from sparse
object boxes to dense voxel-level semantic occupancy, so that persistent
infrastructure, dynamic participants, and free space are evaluated in one
scene-level representation.}

\subsection{Dynamic and Structured Occupancy Modeling}
\label{sec:related_structured}

The closest methods model dynamics or structure within occupancy prediction.
Occupancy flow and 4D forecasting model motion and future
states~\cite{Cam4DOcc,LetOccFlow,UniOcc}, world models learn temporal
evolution~\cite{OccWorld}, and future-instance prediction models BEV
dynamics~\cite{FIERY}. Other lines reweight losses for class
imbalance~\cite{FocalLoss,ClassBalancedLoss} or use sparse-foreground and
coarse-to-fine representations for efficiency~\cite{SparseOcc,COTR}. These
methods model temporal motion, adjust class weights, or reduce computation, but
still assume ego-centric flat voxel classification. \rev{None turns the
fixed-view static-dynamic asymmetry into ordered evidence reasoning.
ProSD-Occ instead reorganizes prediction at the representation and reasoning
level---explaining the stable layout first and recovering dynamics as
residual evidence---rather than through loss or temporal adjustments alone.}

%% file: latex/3-benchmark.tex
\subsection{Benchmark Overview}
\label{sec:benchmark_overview}

\rev{As summarized in Tab.~\ref{tab:dataset_comparison}, existing real-world
datasets support either ego-centric occupancy~\cite{OpenOccupancy,Occ3D} or
object-level roadside/cooperative detection; none provides dense semantic
occupancy under real fixed roadside sensing
(Sec.~\ref{sec:introduction}).}
To fill this gap, we introduce InfraOcc, which constructs dense voxel-level
semantic occupancy annotations in a fixed roadside coordinate system, supports
unified camera-only, LiDAR-only, and multi-modal evaluation, and provides
diagnostic metrics that separate static and dynamic occupancy. Its design goal is
to make the \rev{static-dynamic} asymmetry of roadside scenes \emph{constructible},
\emph{measurable}, and \emph{evaluable}: a \rev{static-dynamic} decoupled annotation
pipeline makes it constructible, temporal and semantic diagnostics make it
measurable, and a static/dynamic evaluation protocol makes it evaluable.

\subsection{Infrastructure-side Occupancy Task}
\label{sec:benchmark_task}

Unlike vehicle-centric occupancy perception, InfraOcc defines the occupancy
field in a fixed roadside coordinate system that remains invariant over time and
is shared by all frames in \rev{the same scene}. The infrastructure-side occupancy task is
formulated as learning a mapping
\begin{equation}
  \mathcal{F}_{\theta}:\ \mathcal{O}_t \;\longmapsto\;
  \mathbf{Y}_t \in (\mathcal{C}\cup\{\mathrm{free}\})^{X\times Y\times Z},
\end{equation}
where the observation
$\mathcal{O}_t\in\{\mathcal{I}_t,\mathcal{P}_t,(\mathcal{I}_t,\mathcal{P}_t)\}$
corresponds to the roadside sensor measurements at timestamp $t$, consisting of
multi-view images $\mathcal{I}_t$, infrastructure-side LiDAR point clouds
$\mathcal{P}_t$, or their combination. The prediction
$\mathbf{Y}_t$ represents a dense semantic occupancy volume over the shared
roadside voxel space, where each voxel is assigned either a semantic occupancy
label from $\mathcal{C}$ \rev{(e.g., car, pedestrian, vegetation, or driveable
surface)} or the free-space label.
\input{latex/fig/sensor-setup.tex}

\subsection{Data Source and Sensor Setup}
\label{sec:benchmark_sensor}
\input{latex/fig/dataset-construction-pipeline.tex}
\input{latex/fig/annotation-verification.tex}

\rev{Our benchmark is derived from real multi-modal sequential streams and
temporally consistent object annotations in the V2XPnP Sequential
Dataset~\cite{V2XPnP}. We select sequences collected around an urban
intersection and its adjacent road segments by two infrastructure units and
associated vehicle agents.} Each infrastructure unit carries one infrastructure-side LiDAR
(Ouster OS1-128) and two \rev{Axis cameras ($1920\times1080$)} under
GPS time synchronization, yielding four calibrated camera views and two
infrastructure-side LiDAR scans per keyframe over a fixed-view camera--LiDAR
system (Fig.~\ref{fig:sensor_setup}). \rev{V2XPnP associates 3D boxes across
time and agents' views with unique tracking IDs. InfraOcc uses these identities
only during offline annotation construction to form tracklets for multi-frame
dynamic-object reconstruction; they are not exposed as benchmark inputs.
We reorganize the original object-level cooperative perception and prediction
setting into a benchmark for dense semantic occupancy prediction. InfraOcc
releases the derived occupancy annotations and toolkit, while the underlying
V2XPnP sensor streams are obtained separately from the official dataset
channel.}
Since all benchmark input sensors are fixed, InfraOcc retains the original calibration and
synchronization to align roadside images, LiDAR point clouds, and occupancy
annotations within one fixed roadside coordinate system, enabling camera-only,
LiDAR-only, and multi-modal evaluation under a shared data split and protocol.
Fig.~\ref{fig:dataset_sample} shows a representative keyframe with the four
roadside camera views together with voxel-level semantic occupancy annotations.

\subsection{Dataset Construction Pipeline}
\label{sec:benchmark_pipeline}

\subsubsection{Pipeline Overview}
\rev{Infrastructure-side occupancy annotation faces two key challenges.}
First, fixed infrastructure-side LiDAR is limited by viewpoint and
occlusion, making it difficult to fully cover persistent infrastructure such as
roads, buildings, vegetation, and barriers. Second, dynamic objects captured by
a single roadside LiDAR scan are often sparse and incomplete, especially around
object boundaries. \rev{To address these challenges, InfraOcc adopts a
semantic point-cloud-based construction strategy that decouples static
background completion from dynamic object reconstruction, as illustrated in
Fig.~\ref{fig:dataset_construction}.} The upper branch constructs
Semantic Dynamic Object Point Clouds from infrastructure-side LiDAR sequences,
while the lower branch constructs Semantic Static Background Point Clouds from
vehicle-side LiDAR sequences. These two semantic point-cloud sources are then
integrated by \rev{Static-Dynamic Recomposition} in the fixed roadside coordinate system
and converted into dense voxel-level semantic occupancy annotations through
Occupancy Labeling. Vehicle-side LiDAR is used only for annotation
construction; the benchmark inputs remain infrastructure-side cameras,
infrastructure-side LiDAR, or their combination according to different
evaluation tracks.

\input{latex/fig/camera-visibility-mask.tex}

\subsubsection{Semantic Dynamic Object Point Clouds}
We first form object tracklets from the source 3D box annotations and unique
tracking IDs provided by V2XPnP~\cite{V2XPnP}. Each dynamic object is thereby
associated with temporally consistent 3D bounding boxes, a semantic category,
and a track identity. Based on these
tracklets, object point clouds belonging to the same dynamic object are cropped
from multiple infrastructure-side LiDAR scans.
To aggregate partial observations of moving objects over time, we perform
Tracklet-based Object Alignment. Specifically, the relative poses of 3D boxes
within the same tracklet are used to estimate a box-guided rigid
transformation, which coarsely aligns object point clouds to the reference
object coordinate system. This step compensates for object motion and avoids
motion ghosting caused by direct accumulation in the global frame. ICP
refinement~\cite{Besl1992ICP} is further applied to reduce local misalignment
caused by tracking jitter, box localization errors, and partial occlusion. The accumulated object
point clouds inherit the semantic category of the corresponding tracklet,
forming denser and semantically consistent dynamic object geometry.

\subsubsection{Semantic Static Background Point Clouds}
Since vehicle-side LiDAR observes the same
infrastructure scene from moving viewpoints, it provides more complete
geometric coverage for persistent static elements. Multi-frame vehicle-side point
clouds are first registered into the roadside coordinate system through
Ego-pose Alignment. Dynamic object points are then removed to prevent transient
traffic participants from being fused into the static background.
Semantic Annotation is performed on the accumulated static background to assign
semantic labels to persistent infrastructure elements. The resulting Semantic
Static Background Point Clouds provide a dense static scene scaffold for the
subsequent \rev{Static-Dynamic Recomposition}.

\subsubsection{Static-Dynamic Recomposition}
After constructing Semantic Dynamic Object Point Clouds and Semantic Static
Background Point Clouds, we recompose them into keyframe-level semantic scenes
in the fixed roadside coordinate system.
For each keyframe, the Semantic Static Background Point Clouds serve as a
persistent scene scaffold. Dynamic object point clouds are transformed into the
roadside coordinate system according to their tracklet poses at the current
timestamp and inserted into the static background. This produces a semantic 3D
scene that preserves both stable infrastructure layout and keyframe-specific
traffic participants, while avoiding motion artifacts caused by directly
accumulating moving objects.

\noindent\textbf{Image-guided Verification.}
We further project the recomposed semantic scene onto calibrated roadside
camera views to inspect its consistency with image observations. As shown in
Fig.~\ref{fig:2d_3d_consistency}, this verification checks semantic boundaries,
dynamic object locations, and cross-modal calibration consistency before
voxel-level occupancy labeling.

\subsubsection{Occupancy Labeling}
After static-dynamic recomposition, each keyframe is represented as a semantic
3D scene containing both persistent infrastructure layout and keyframe-specific
dynamic objects. Occupancy Labeling converts this recomposed scene into
voxel-level semantic occupancy ground truth. Similar to Occ3D~\cite{Occ3D}, we
distinguish three voxel states: occupied, free, and unobserved, so that empty
space and occluded regions are not confused.

\noindent\textbf{Semantic Occupancy GT.}
We first discretize the recomposed semantic point clouds into the predefined
voxel grid. Voxels containing semantic points are labeled as occupied and
assigned the corresponding semantic category. Static background points provide
labels for persistent infrastructure elements, while dynamic object points
provide labels for traffic participants at the current keyframe. In this way,
static layout and dynamic objects are represented in the same fixed roadside
coordinate system.

\noindent\textbf{Visibility Reasoning.}
Voxels without semantic points are further processed by visibility reasoning.
We trace infrastructure-side LiDAR rays from the sensor origin to observed
surfaces. Voxels traversed before the first occupied surface are labeled as
free space, while voxels behind observed surfaces or outside valid sensor rays
are treated as unobserved and ignored. This prevents occluded regions behind
buildings, vegetation, or vehicles from being incorrectly labeled as free.
We further compute camera-view visibility masks by projecting voxel centers
into calibrated roadside cameras. A voxel is regarded as camera-visible if its
projected center lies inside the image boundary with positive depth.
Fig.~\ref{fig:camera_mask} visualizes the resulting masks, which characterize
the image-observable regions of different roadside cameras. \rev{Following
Occ3D~\cite{Occ3D}, these masks are released with the annotations to support
optional camera-visible evaluation, while the main protocol evaluates the
full grid (Sec.~\ref{sec:benchmark_eval}).}

\subsection{Dataset Statistics}
\label{sec:benchmark_statistics}

\subsubsection{Scale and Annotation Coverage}
InfraOcc contains 290 temporally continuous roadside sequences (215 for
training and 75 for \rev{testing}), each spanning roughly 10--24 seconds. Raw
streams are recorded at 10 Hz, while dense semantic occupancy is annotated on
2 Hz keyframes (every fifth raw frame). Each annotated
keyframe provides four calibrated roadside camera views, two infrastructure-side
LiDAR scans, and one dense voxel-level occupancy annotation over the range
$[-64,64]\!\times\![-64,64]\!\times\![-4.8,1.6]$~m at a $0.4$~m voxel size, i.e.,
a $320\times320\times16$ grid; its label space comprises the occupied semantic
classes and a free-space class, supporting both semantic and category-agnostic
geometry evaluation.

\input{latex/fig/temporal-persistence-statistics.tex}
\input{latex/fig/class-distribution.tex}

\subsubsection{Static-Dynamic Temporal Asymmetry}
\rev{To quantify the temporal structure induced by fixed infrastructure-side
observation, we measure, within each scene, how persistently each BEV cell is
occupied. For each BEV cell, the \emph{occupied-frame ratio} is the number
of keyframes in which the cell contains static (or dynamic) voxels, divided
by the total number of keyframes in the scene: a ratio of 100\% means the
location is occupied all the time, whereas a low ratio means it is occupied
only transiently. For
visualization,
semantic labels are
regrouped into coarse categories, while benchmark evaluation still follows the
full semantic label space.
As shown in Fig.~\ref{fig:spatiotemporal_semantic_statistics}(a) and (b),
static groups form spatially continuous, high-persistence layouts, whereas
dynamic occupancy appears only as sparse, low-intensity traces along lanes and
crosswalks: dynamic objects repeatedly pass through certain regions but rarely
occupy the same BEV cell for long. The persistence curve in
Fig.~\ref{fig:spatiotemporal_semantic_statistics}(c) quantifies this contrast:
static cells remain almost fully retained across most thresholds, with a
median occupied-frame ratio of 100\%, whereas dynamic cells quickly vanish,
with a median ratio of only 1.8\% and a 95th percentile of 16.5\%. These
observations confirm the persistent-static versus transient-dynamic structure
of infrastructure-side occupancy.}

\subsubsection{Semantic Long-tail Distribution}
In addition to temporal asymmetry, InfraOcc also exhibits a pronounced semantic
frequency imbalance. Fig.~\ref{fig:semantic_class_distribution} reports the
class-wise voxel occupancy ratio over evaluated occupied classes, excluding
free space and ignored categories. Static infrastructure categories account for
97.3\% of occupied voxels, whereas dynamic traffic participants account for
only 2.7\%.
\rev{Within each group the distribution is further skewed: driveable surface
and manmade structures dominate the static classes, while cars dominate the
dynamic classes, with trucks, motorcycles, and bicycles substantially
sparser.} This semantic long-tail, together with the temporal
asymmetry above, suggests that infrastructure-side occupancy prediction requires
not only overall semantic accuracy, but also group-wise diagnosis of persistent
static layout and sparse dynamic objects.

\subsection{Evaluation Protocol}
\label{sec:benchmark_eval}

InfraOcc evaluates camera-only, LiDAR-only, and multi-modal occupancy under
a unified protocol: all tracks share the same data split, label space,
calibration, and fixed roadside coordinate system, and are evaluated over the
full $320\times320\times16$ grid (Sec.~\ref{sec:benchmark_statistics}) rather
than only sensor-visible regions\rev{; camera-visibility masks
(Sec.~\ref{sec:benchmark_pipeline}) are additionally provided for optional
camera-visible evaluation}. Following common occupancy
benchmarks~\cite{OpenOccupancy,Occ3D}, the primary metric is the mean IoU over
valid occupied classes,
\begin{equation}
  \mathrm{mIoU}_{\mathrm{all}}
  =
  \frac{1}{|\mathcal{C}_{\mathrm{eval}}|}
  \sum_{c\in\mathcal{C}_{\mathrm{eval}}}
  \frac{\mathrm{TP}_{c}}{\mathrm{TP}_{c}+\mathrm{FP}_{c}+\mathrm{FN}_{c}},
\end{equation}
where $\mathcal{C}_{\mathrm{eval}}$ excludes free space and the ignored
categories (construction vehicle, trailer, and other flat; omitted due to
sparsity or ambiguous boundaries), and free space is reported separately as
\textit{Free} IoU. To diagnose the \rev{static-dynamic} asymmetry, we additionally
report gIoU (the IoU of all non-free classes merged into a single occupied
group) and the group-wise mIoU$_{\mathrm{dyn}}$ / mIoU$_{\mathrm{sta}}$, i.e.,
mIoU restricted to the dynamic classes (bicycle, bus, car, motorcycle,
pedestrian, truck) and the static classes (others, barrier, traffic cone,
driveable surface, sidewalk, terrain, manmade, vegetation), respectively.
Together with per-class IoU, these metrics evaluate geometry, static layout,
dynamic objects, and free space across modalities.

%% file: latex/fig/sensor-setup.tex
\begin{figure}[t]
  \centering
  \includegraphics[width=\linewidth]{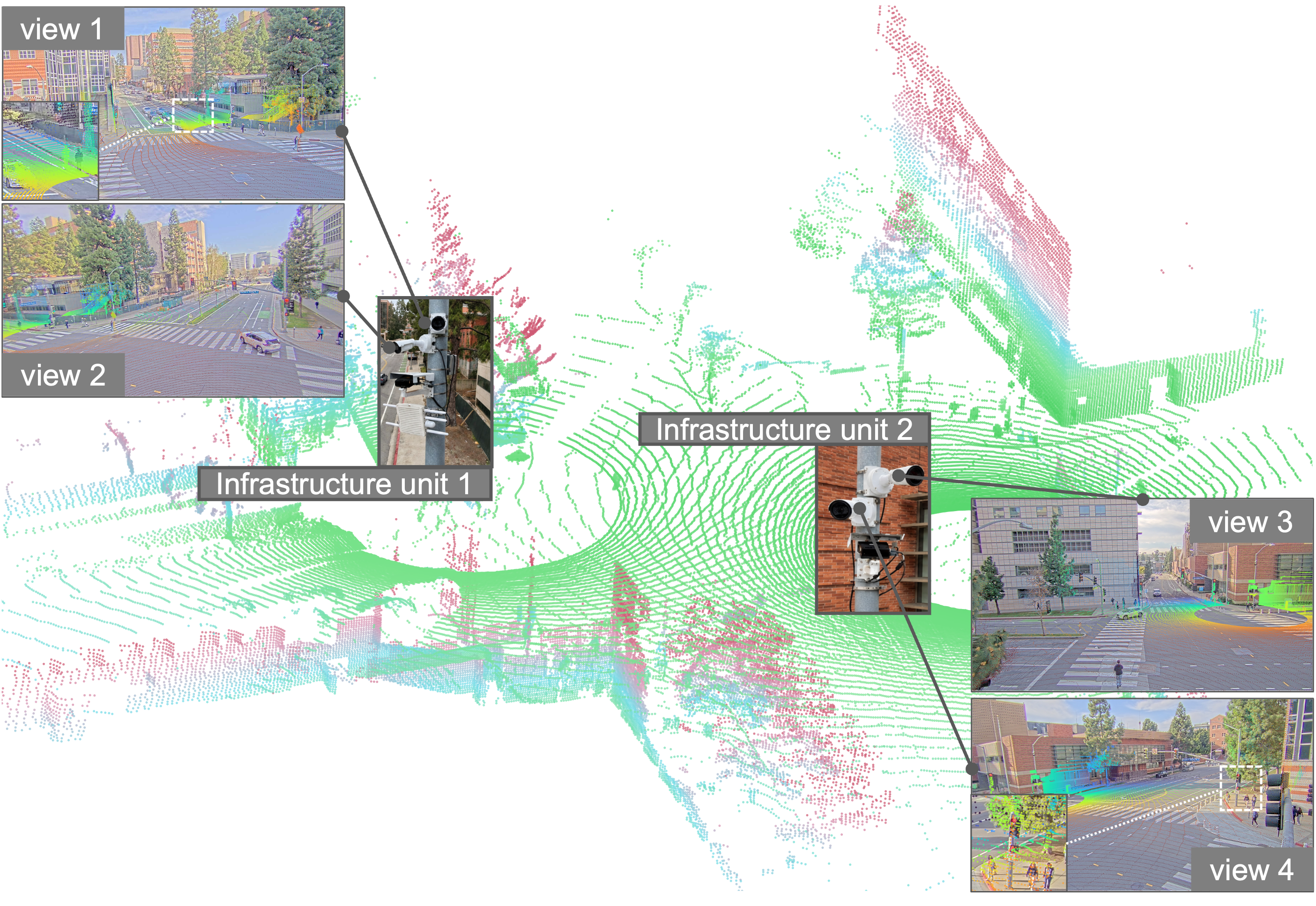}
  \caption{\textbf{Infrastructure-side sensor configuration.}
  The roadside platform comprises two fixed infrastructure units, each equipped
  with one LiDAR and two high-resolution cameras under GPS time synchronization.
  Their calibrated streams are organized in a fixed roadside coordinate system
  and support camera-only, LiDAR-only, and multi-modal occupancy prediction.}
  \label{fig:sensor_setup}
\end{figure}

%% file: latex/fig/dataset-construction-pipeline.tex
\begin{figure*}[!t]
  \centering
  \includegraphics[width=\textwidth]{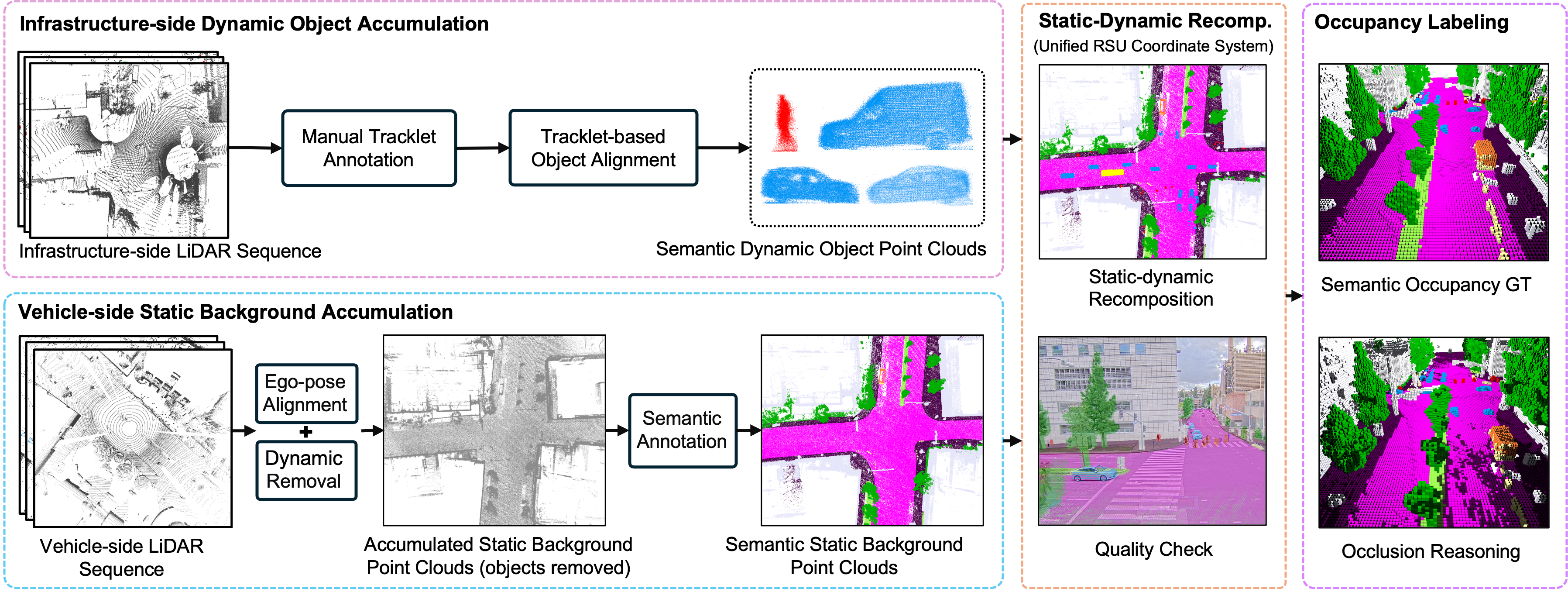}
\caption{\textbf{Overview of the dataset construction pipeline.} We build dense semantic occupancy annotations from two complementary
semantic point-cloud sources: Semantic Dynamic Object Point Clouds constructed from infrastructure-side LiDAR sequences and Semantic Static Background Point
Clouds constructed from vehicle-side LiDAR sequences. \rev{The dynamic branch forms object tracklets from source 3D box annotations and
tracking IDs, and then uses Tracklet-based Object Alignment to densify dynamic objects, while the static
branch uses Ego-pose Alignment, Dynamic-Object Removal, and Semantic
Annotation to construct persistent infrastructure layout. Static-Dynamic
Recomposition integrates the two semantic point-cloud sources in the fixed
roadside coordinate system, and Occupancy Labeling converts the recomposed
scene into voxel-level semantic occupancy labels.}}
\label{fig:dataset_construction}
\end{figure*}

%% file: latex/fig/annotation-verification.tex
\begin{figure}[!t]
  \centering
  \includegraphics[width=\linewidth]{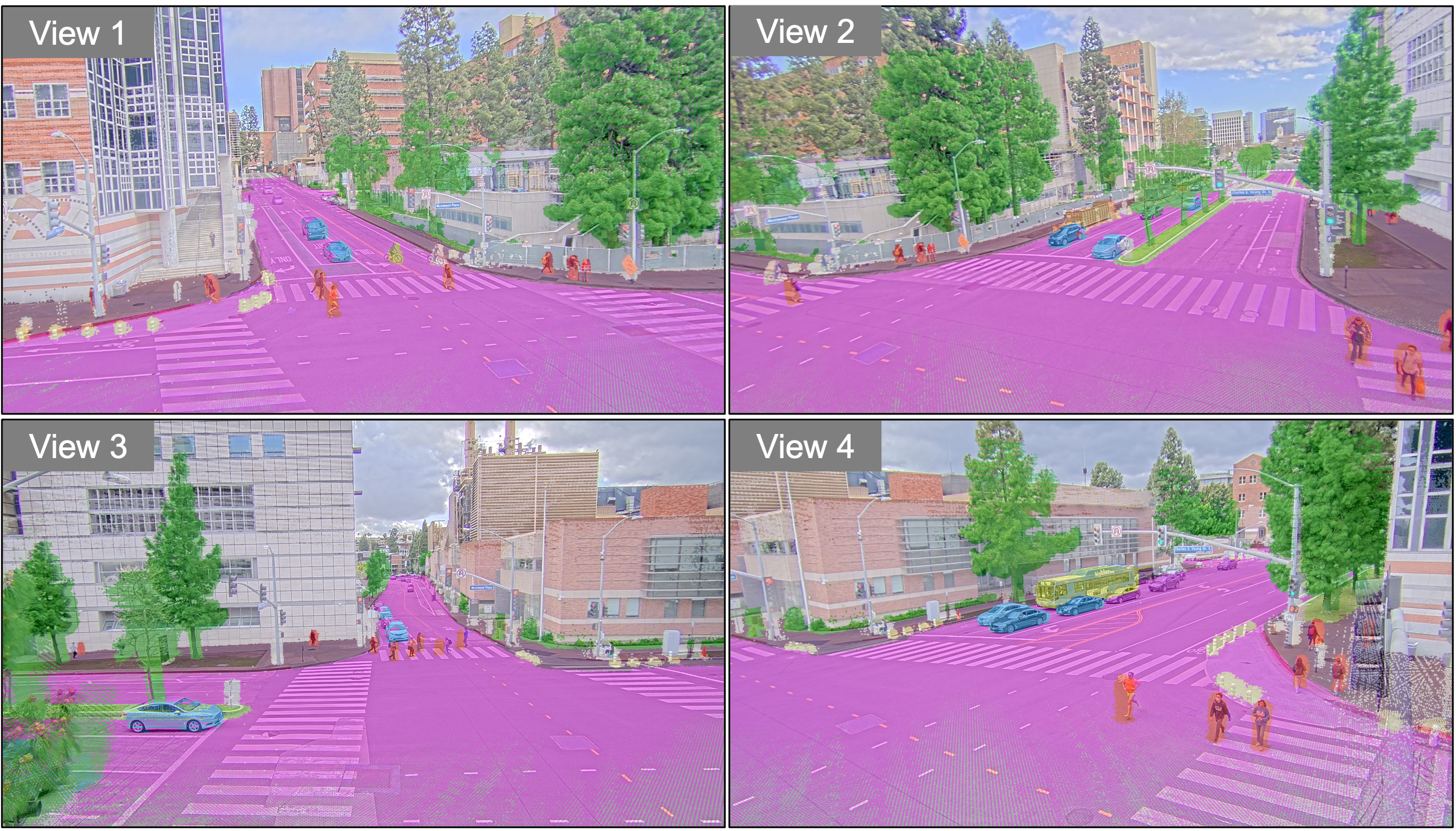}
  \caption{\textbf{\rev{Image-guided annotation verification.}} The recomposed semantic 3D scene is projected onto the calibrated roadside
  camera views to inspect the alignment between dense point-cloud semantic
  annotations and image observations. Annotators check semantic boundaries,
  object placement, and cross-modal calibration consistency, and manually correct
  any misclassified labels, thereby ensuring the reliability of the
  constructed infrastructure-side occupancy annotations.}
  \label{fig:2d_3d_consistency}
  \vspace{-3pt}
\end{figure}

%% file: latex/fig/camera-visibility-mask.tex
\begin{figure*}[!t]
  \centering
  \includegraphics[width=\textwidth]{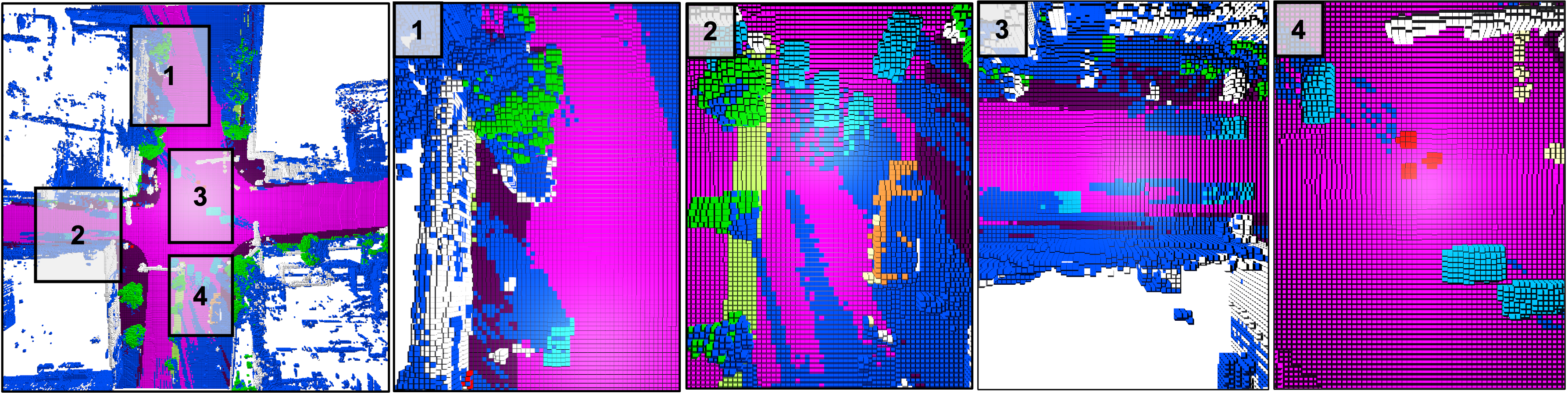}
  \caption{\textbf{Camera-view visibility masks for occupancy labeling.}
  The voxel grid is projected into calibrated roadside camera views to identify
  image-observable regions in the fixed roadside coordinate system. The left
  panel overlays the visibility masks of four roadside views on the semantic
  occupancy map, while the right panels show enlarged local regions for each
  camera view. \rev{These masks support consistency inspection between
  roadside images and voxel-level semantic occupancy annotations.}}
  \label{fig:camera_mask}
\end{figure*}

%% file: latex/fig/temporal-persistence-statistics.tex
\begin{figure}[t]
  \centering
  \includegraphics[width=\linewidth]{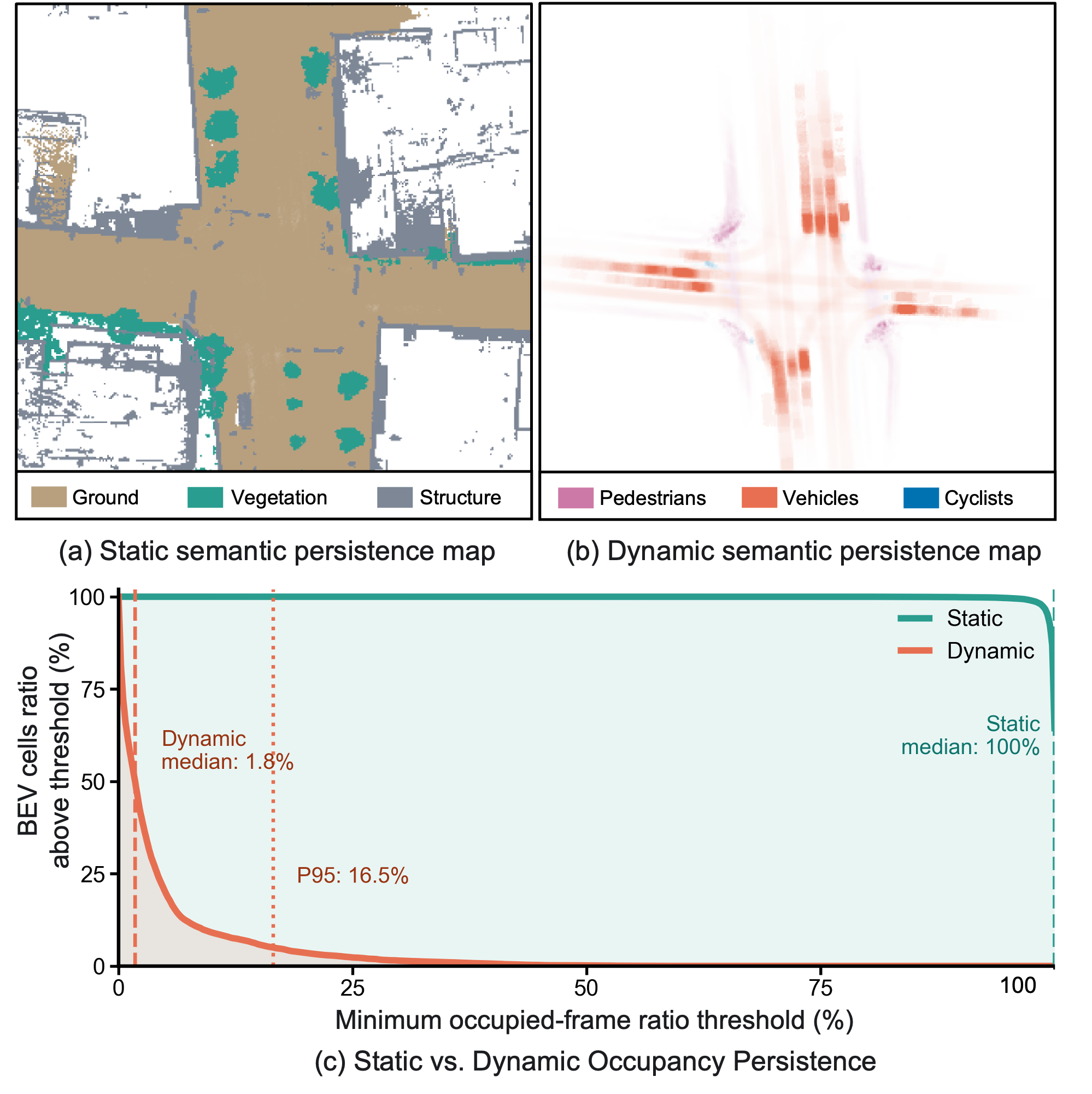}
  \caption{\textbf{Static-dynamic occupancy persistence statistics.}
(a) Static semantic persistence map in the fixed infrastructure coordinate
system. Colors denote dominant coarse static groups, and color intensity
indicates the occupied-frame ratio of each BEV cell. (b) Dynamic semantic
transience map computed in the same BEV space, where dynamic occupancy mainly
appears along lanes, crosswalks, and typical traffic trajectories. (c)
Occupancy persistence curve. \rev{The horizontal axis denotes the minimum
occupied-frame ratio threshold, and the vertical axis denotes the ratio of
BEV cells retained above the threshold. Static regions remain highly
persistent, whereas dynamic regions are sparse and transient.}
}
\label{fig:spatiotemporal_semantic_statistics}
\end{figure}

%% file: latex/fig/class-distribution.tex
\begin{figure}[h!t]
  \centering
  \includegraphics[width=\linewidth]{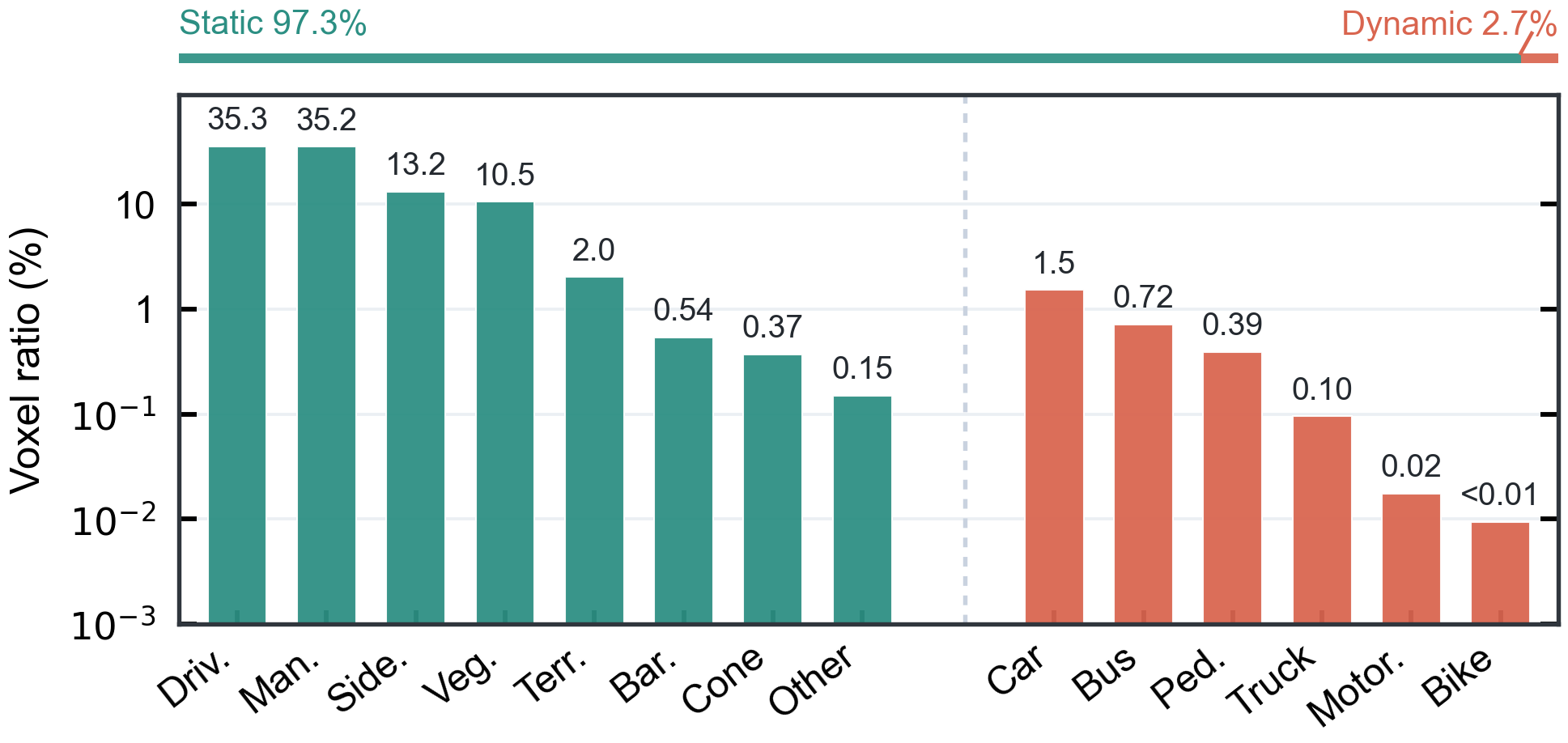}
  \caption{\textbf{Semantic class distribution.}
The top strip summarizes occupied voxels by semantic group, while the lower
panel reports class-wise voxel occupancy ratios over evaluated occupied
classes. The vertical axis is
shown in log scale for readability. The distribution shows that static
infrastructure dominates the occupied semantic space, whereas dynamic traffic
participants form a sparse long-tailed subset.}
  \label{fig:semantic_class_distribution}
\end{figure}

%% file: latex/table_macros.tex
\newcommand{\bestnum}[1]{\textbf{#1}}
\newcommand{\secondnum}[1]{\underline{#1}}
\newcommand{\thirdnum}[1]{\emph{#1}}
\newcolumntype{C}[1]{>{\centering\arraybackslash}p{#1}}
\newcommand{\bestcell}[1]{\textbf{#1}}
\newcommand{\secondcell}[1]{\underline{#1}}
\newcommand{\thirdcell}[1]{\emph{#1}}
\newcommand{\rotcls}[1]{\rotatebox{90}{\scriptsize #1}}

\definecolor{occBike}{RGB}{244,160,170}
\definecolor{occBus}{RGB}{241,120,52}
\definecolor{occCar}{RGB}{31,144,224}
\definecolor{occMotor}{RGB}{188,165,0}
\definecolor{occPed}{RGB}{226,35,35}
\definecolor{occTruck}{RGB}{153,78,24}
\definecolor{occOther}{RGB}{125,125,125}
\definecolor{occBarrier}{RGB}{96,62,45}
\definecolor{occCone}{RGB}{245,213,93}
\definecolor{occDrive}{RGB}{230,30,225}
\definecolor{occSide}{RGB}{85,21,91}
\definecolor{occTerrain}{RGB}{125,224,80}
\definecolor{occManmade}{RGB}{196,196,196}
\definecolor{occVegetation}{RGB}{0,185,0}
\definecolor{occFree}{RGB}{0,0,0}

\newcommand{\classswatch}[1]{\textcolor{#1}{\rule{0.70em}{0.70em}}}
\newcommand{\occclassname}[1]{\rotatebox{90}{\scriptsize #1}}
\newcommand{\occclasshead}[2]{%
  \begin{tabular}[b]{@{}c@{}}%
    \makebox[0.85em][c]{\rule{0pt}{6.4em}\rotatebox{90}{\scriptsize #2}}\\[-0.25ex]%
    \classswatch{#1}%
  \end{tabular}%
}
\newcommand{\modalityrow}[2]{%
  \rowcolor{gray!8}\multicolumn{#1}{@{}l}{\textbf{#2}}\\%
}

%% file: latex/3-method.tex
\subsection{Motivation}
\label{sec:prosd_design_motivation}
As established in Sec.~\ref{sec:introduction} and the benchmark diagnostics
(Sec.~\ref{sec:benchmark_statistics}), the fundamental difficulty of
fixed-viewpoint roadside occupancy is the structural asymmetry between
persistent static layout and transient dynamic participants. Under this
structure, flat one-shot voxel classification is biased toward spatially
dominant static patterns and weakens sparse dynamic cues, while simple loss
reweighting~\cite{FocalLoss,ClassBalancedLoss} only adjusts class weights
without changing how static and dynamic evidence are organized. Accordingly, the
design principle of ProSD-Occ is to organize occupancy reasoning from stable
layout to local residual evidence rather than predicting all voxel categories at
once: the model first establishes a sample-adaptive explanation of the
persistent static layout, then uses it as a reference to recover residual
evidence that the static context cannot sufficiently explain---often
corresponding to sparse yet safety-critical dynamic objects---and finally
recomposes static, dynamic, and free-space evidence into a unified semantic
occupancy field. This paradigm exploits the contextual value of stable layout
while suppressing its dominance over sparse dynamic prediction.
\input{latex/fig/method-overview}

\subsection{Overall Architecture}
\label{sec:prosd_overview}

As shown in \Cref{fig:method_overview_refined}, ProSD-Occ organizes
infrastructure-side semantic occupancy prediction as a progressive
static-to-dynamic reasoning framework. The framework decouples modality-specific
feature construction from modality-agnostic occupancy reasoning. Given
camera-only, LiDAR-only, or multi-modal inputs, Modality-flexible Feature
Encoding first converts the observations into a unified voxel representation
$\bm{F}$. In the camera-only and LiDAR-only settings, $\bm{F}$ is instantiated
by the corresponding camera or LiDAR voxel feature. In the multi-modal setting,
camera and LiDAR voxel features are fused before entering the subsequent
reasoning process. This unified representation enables the same
static-to-dynamic reasoning pipeline to be applied across all sensing regimes.

Given $\bm{F}$, ProSD-Occ proceeds from stable layout explanation to dynamic
evidence recovery. Static Layout Reasoner first estimates static logits,
probabilities, and confidence $(\bm{Z}_{s},\bm{P}_{s},\bm{C}_{s})$, where
$\bm{C}_{s}$ provides a sample-adaptive soft explanation of persistent
infrastructure layout. Static-guided Residual Modulation then uses
$\bm{C}_{s}$ to attenuate static-dominant responses while preserving
complementary residual information, producing the dynamic-aware feature
$\bm{F}_{d}$. Based on $\bm{F}_{d}$, Residual Dynamic Predictor estimates
dynamic logits, probabilities, and confidence
$(\bm{Z}_{d},\bm{P}_{d},\bm{C}_{d})$, focusing on local and transient object
evidence. Finally, Semantic Recomposition integrates static, dynamic, and
free-space evidence at the logit level and outputs the final occupancy logits
$\bm{Z}_{\mathrm{final}}$.

\subsection{Modality-flexible Feature Encoding}
\label{sec:prosd_modality_flexible}
ProSD-Occ performs static-to-dynamic occupancy reasoning on a unified voxel
representation, regardless of the input sensing modality. Given camera images,
LiDAR points, or both, the modality-specific front-end first maps the
observations into a shared 3D voxel feature tensor
\begin{equation}
  \bm{F}\in\mathbb{R}^{C_f\times Z\times H\times W},
\end{equation}
which serves as the common input to the subsequent Progressive
Static-to-Dynamic Reasoning module.

For camera input, we follow common image-to-BEV/voxel encoding
designs~\cite{BEVDet,BEVDepth}. Multi-view images are first processed by an
image backbone and neck, and then lifted into the 3D voxel space through a
depth-aware view transformation. The resulting camera voxel features are
further refined with a BEVFormer-style backward projection module~\cite{BEVFormer}
to aggregate multi-view geometric context. We denote the resulting camera voxel
feature as $\bm{F}_{\mathrm{cam}}$.
For LiDAR input, point clouds are converted into voxel representations following
standard voxel-based point cloud encoding~\cite{VoxelNet,PointPillars}. The
voxelized points are encoded by a sparse LiDAR encoder and then processed by a
3D convolutional backbone to obtain LiDAR voxel features, denoted as
$\bm{F}_{\mathrm{lidar}}$.
In the multi-modal setting, we follow the unified BEV/voxel fusion paradigm in
multi-sensor perception~\cite{BEVFusion,OpenOccupancy}. Camera and LiDAR
features are first projected into the same voxel grid and then fused before
occupancy reasoning. Concretely, the fusion block computes lightweight
attention maps from each modality and uses them to cross-modulate the voxel
features of the other modality. The modulated camera and LiDAR features are
concatenated and projected back to the shared channel dimension by a 3D
convolutional block, producing the fused voxel feature
$\Phi(\bm{F}_{\mathrm{cam}},\bm{F}_{\mathrm{lidar}})$.

\rev{The input to \nameplus{} is then uniformly defined as}
\begin{equation}
  \bm{F} =
  \begin{cases}
  \bm{F}_{\mathrm{cam}}, & \text{Camera-only},\\
  \bm{F}_{\mathrm{lidar}}, & \text{LiDAR-only},\\
  \Phi(\bm{F}_{\mathrm{cam}},\bm{F}_{\mathrm{lidar}}), & \text{Multi-modal}.
  \end{cases}
\end{equation}

\subsection{Progressive Static-to-Dynamic Reasoning}
\label{sec:prosd_reasoning}
\rev{We now detail the four components of the progressive reasoning module
(\Cref{fig:method_overview_refined}).}
\subsubsection{Static Layout Reasoner}
\label{sec:prosd_static_reasoner}
This module estimates persistent static layout from the unified voxel feature.
Its goal is to explain stable infrastructure elements before dynamic reasoning
is performed. Let $\mathcal{S}$ denote the static class set
\rev{(Sec.~\ref{sec:benchmark_eval})}. \rev{As shown in
\Cref{fig:method_overview_refined}, the Static Layout Reasoner is implemented
by a compact residual 3D convolutional tower (BasicBlock3D) followed by a
voxel-wise two-layer MLP, mapping each voxel feature to logits over
$\mathcal{S}$ and an additional ``other'' class:}
\begin{equation}
  \bm{Z}_{s}=h_s(\bm{F}),
  \quad
  \bm{Z}_{s}\in
  \mathbb{R}^{(|\mathcal{S}|+1)\times Z\times H\times W} .
\end{equation}

Different from a generic background category, the ``other'' class in the static
branch explicitly represents voxels that are not explained by static layout,
including dynamic objects and free space. Therefore, the static branch is not
asked to solve the full semantic occupancy problem. Instead, it only estimates
whether each voxel can be explained by persistent static structure. The static
probability is computed by
\begin{equation}
  \bm{P}_{s}=\operatorname{softmax}(\bm{Z}_{s}).
\end{equation}
We define the static confidence as the probability that a voxel does not belong
to the static-branch ``other'' class:
\begin{equation}
  \bm{C}_{s}
  =
  1-\bm{P}_{s}(\mathrm{other}).
\end{equation}
The confidence $\bm{C}_{s}$ provides a sample-adaptive soft explanation of
persistent layout. It is not used as a hard static mask; instead, it guides the
subsequent residual modulation step, where static-dominant responses are
suppressed to expose potential dynamic evidence.

\subsubsection{Static-guided Residual Modulation}
\label{sec:prosd_residual_modulation}

Static-guided Residual Modulation bridges static layout reasoning and residual
dynamic prediction by transforming the unified voxel feature $\bm{F}$ into a
dynamic-aware feature $\bm{F}_{d}$. As shown in
\Cref{fig:method_overview_refined}, this module contains two complementary
paths: a suppression path that attenuates static-dominant responses according
to the predicted static confidence, and a residual path that preserves raw voxel
evidence to avoid over-suppression.

Given the static confidence $\bm{C}_{s}$, the suppression path constructs a
voxel-wise suppression gate:
\begin{equation}
  \alpha=\sigma(a),
  \quad
  \bm{G}
  =
  \operatorname{clamp}
  \left(1-\alpha\,\operatorname{sg}(\bm{C}_{s}),0,1\right),
\end{equation}
where $a$ is a learnable scalar, $\alpha\in(0,1)$ controls the suppression
strength, and $\operatorname{sg}(\cdot)$ denotes stop-gradient. Voxels with high
static confidence are softly attenuated, encouraging the subsequent dynamic
branch to focus on evidence not sufficiently explained by static layout. The
static-suppressed feature is obtained by
\begin{equation}
  \bm{F}_{\mathrm{sup}}=\bm{F}\odot\bm{G}.
\end{equation}

Since static confidence can be uncertain around object boundaries, occluded
regions, or sparse dynamic objects, pure suppression may discard useful dynamic
cues. The residual path therefore introduces a bypass from the original voxel
feature:
\begin{equation}
  \bm{F}_{\mathrm{raw}}
  =
  \operatorname{BN}
  \left(\operatorname{Conv}_{1\times1\times1}(\bm{F})\right),
  \quad
  \beta=\sigma(b),
\end{equation}
where $b$ is a learnable scalar and $\beta\in(0,1)$ controls the bypass
strength. The final dynamic-aware feature is then formed as
\begin{equation}
  \bm{F}_{d}
  =
  \bm{F}_{\mathrm{sup}}+\beta\bm{F}_{\mathrm{raw}}.
\end{equation}
In this way, the module suppresses dominant static responses while retaining
complementary residual evidence, providing a more suitable representation for
dynamic occupancy prediction.

\subsubsection{Residual Dynamic Predictor}
\label{sec:prosd_dynamic_predictor}
The Residual Dynamic Predictor estimates transient traffic participants from
the dynamic-aware feature $\bm{F}_{d}$. Different from the Static Layout
Reasoner, this module focuses on local, sparse, and boundary-sensitive object
evidence that remains after static-guided modulation. It adopts the same
compact BasicBlock3D tower and voxel-wise MLP predictor, but predicts over the
dynamic class set $\mathcal{D}$ and an additional ``other'' class:
\begin{equation}
  \bm{Z}_{d}=h_d(\bm{F}_{d}),
  \quad
  \bm{Z}_{d}\in
  \mathbb{R}^{(|\mathcal{D}|+1)\times Z\times H\times W} .
\end{equation}
\rev{Symmetric to the static branch, the dynamic-branch ``other'' class
covers voxels not explained by dynamic objects.} The dynamic
probability and confidence are computed as
\begin{equation}
  \bm{P}_{d}=\operatorname{softmax}(\bm{Z}_{d}),
  \quad
  \bm{C}_{d}=1-\bm{P}_{d}(\mathrm{other}).
\end{equation}
Since $\bm{F}_{d}$ has already been modulated by static confidence, this
branch is encouraged to recover residual dynamic evidence rather than relearn
dominant static layout from scratch.

% Queue the main double-column table before Sec. 5 so it can appear earlier.

\subsubsection{Semantic Recomposition}
\label{sec:prosd_semantic_recomposition}
\rev{After static and dynamic reasoning, the two branches provide
complementary partial explanations, but neither is defined over the complete
semantic label space, and naive averaging would blur their specialization. We
instead recompose the branch outputs at the logit level.}
Specifically, given the static outputs $(\bm{P}_{s},\bm{C}_{s})$ and dynamic
outputs $(\bm{P}_{d},\bm{C}_{d})$, we construct a voxel-wise recomposition
input:
\begin{equation}
  \bm{x}_{g}
  =
  [\bm{P}_{s},\bm{P}_{d},\bm{C}_{s},\bm{C}_{d}].
\end{equation}
A lightweight MLP followed by a sigmoid function predicts three group-wise
recomposition weights:
\begin{equation}
  [\bm{w}_{\mathrm{static}},\bm{w}_{\mathrm{dynamic}},\bm{w}_{\mathrm{free}}]
  =
  \sigma(g_{\phi}(\bm{x}_{g})).
\end{equation}
Here, $\bm{w}_{\mathrm{static}}$ and $\bm{w}_{\mathrm{dynamic}}$ measure
reliance on their corresponding specialized branch, while
$\bm{w}_{\mathrm{free}}$ interpolates the two branch-specific ``other'' logits.

The branch logits $\bm{Z}_{s}$ and $\bm{Z}_{d}$ are then recomposed according to
their semantic groups. Let $\bm{Z}_{s,o}$ and $\bm{Z}_{d,o}$ denote the
``other'' logits of the static and dynamic branches. For a static class
$c\in\mathcal{S}$, the final logit is composed from the corresponding static
logit and the dynamic-branch ``other'' logit:
\begin{equation}
  \bm{Z}_{\mathrm{final},c}
  =
  \bm{w}_{\mathrm{static}}\bm{Z}_{s,c}
  +(1-\bm{w}_{\mathrm{static}})\bm{Z}_{d,o}.
\end{equation}
For a dynamic class $c\in\mathcal{D}$, the final logit is composed from the
dynamic-branch class logit and the static-branch ``other'' logit:
\begin{equation}
  \bm{Z}_{\mathrm{final},c}
  =
  \bm{w}_{\mathrm{dynamic}}\bm{Z}_{d,c}
  +(1-\bm{w}_{\mathrm{dynamic}})\bm{Z}_{s,o}.
\end{equation}
For the free-space class $f$, the final logit is obtained by combining the
``other'' logits from both branches:
\begin{equation}
  \bm{Z}_{\mathrm{final},f}
  =
  \bm{w}_{\mathrm{free}}\bm{Z}_{s,o}
  +(1-\bm{w}_{\mathrm{free}})\bm{Z}_{d,o}.
\end{equation}

\input{latex/tab/main-results}

% Queue the main qualitative figure before the ablation floats.

\subsection{Optimization Objectives}
\label{sec:prosd_objective}

We supervise three outputs in ProSD-Occ: the final occupancy prediction and the
two intermediate static and dynamic branch predictions. The final logits
$\bm{Z}_{\mathrm{final}}$ are supervised in the complete semantic occupancy
space, yielding $\mathcal{L}_{\mathrm{occ}}$. For the static and dynamic
branches, semantic classes outside the corresponding group are mapped to the
branch-specific ``other'' class, yielding $\mathcal{L}_{s}$ and
$\mathcal{L}_{d}$. All three losses adopt the same composite occupancy
objective:
\begin{equation}
  \mathcal{L}_{r}
  =
  \mathcal{L}_{r}^{\mathrm{CE}}
  +
  \mathcal{L}_{r}^{\mathrm{Sem}}
  +
  \mathcal{L}_{r}^{\mathrm{Lovasz}},
  \quad r\in\{\mathrm{occ},s,d\},
\end{equation}
where $\mathcal{L}^{\mathrm{CE}}$, $\mathcal{L}^{\mathrm{Sem}}$, and
$\mathcal{L}^{\mathrm{Lovasz}}$ denote cross-entropy loss, semantic scaling
loss~\cite{MonoScene}, and Lovasz loss~\cite{Berman_2018_CVPR}, respectively.

Finally, since $\bm{C}_{s}$ directly controls the static-guided suppression
gate, its stability matters: persistent structures observed from a fixed
viewpoint should yield consistent static confidence across same-scene frames,
yet per-sample supervision alone leaves it fluctuating under transient
occlusions, dynamic objects, and sensing noise. We therefore add a
\rev{static-confidence consistency regularizer on confidently static voxels.
Because all frames of a scene share the fixed roadside coordinate system,
their static confidences are voxel-aligned and directly comparable once
spatial augmentation is undone. Within each same-scene minibatch group, a
voxel $v$ is treated as confidently static for sample $i$ if
$\bm{C}_{s,i}(v)>\tau$. On voxels where at least two samples are confident,
the consensus $\bar{\bm{C}}_{s}(v)$ is defined as the mean confidence of
these samples, and the regularizer penalizes their squared deviations,
\begin{equation}
  \mathcal{L}_{\mathrm{cons}}
  =
  \operatorname{mean}_{(i,v)}
  \left(
  \bm{C}_{s,i}(v)-\bar{\bm{C}}_{s}(v)
  \right)^2 ,
\end{equation}
averaged over all such sample--voxel pairs. Since the regularizer acts only
on confidently static voxels and the confidence itself is anchored by the
static-branch supervision $\mathcal{L}_{s}$, it removes per-frame
fluctuations on persistent layout without collapsing the confidence field.}

The overall objective is
\begin{equation}
  \mathcal{L}
  =
  \mathcal{L}_{\mathrm{occ}}
  +
  \lambda_s\mathcal{L}_{s}
  +
  \lambda_d\mathcal{L}_{d}
  +
  \lambda_c\mathcal{L}_{\mathrm{cons}}.
\end{equation}

%% file: latex/fig/method-overview.tex
\begin{figure*}[!t]
\centering
\includegraphics[width=\textwidth]{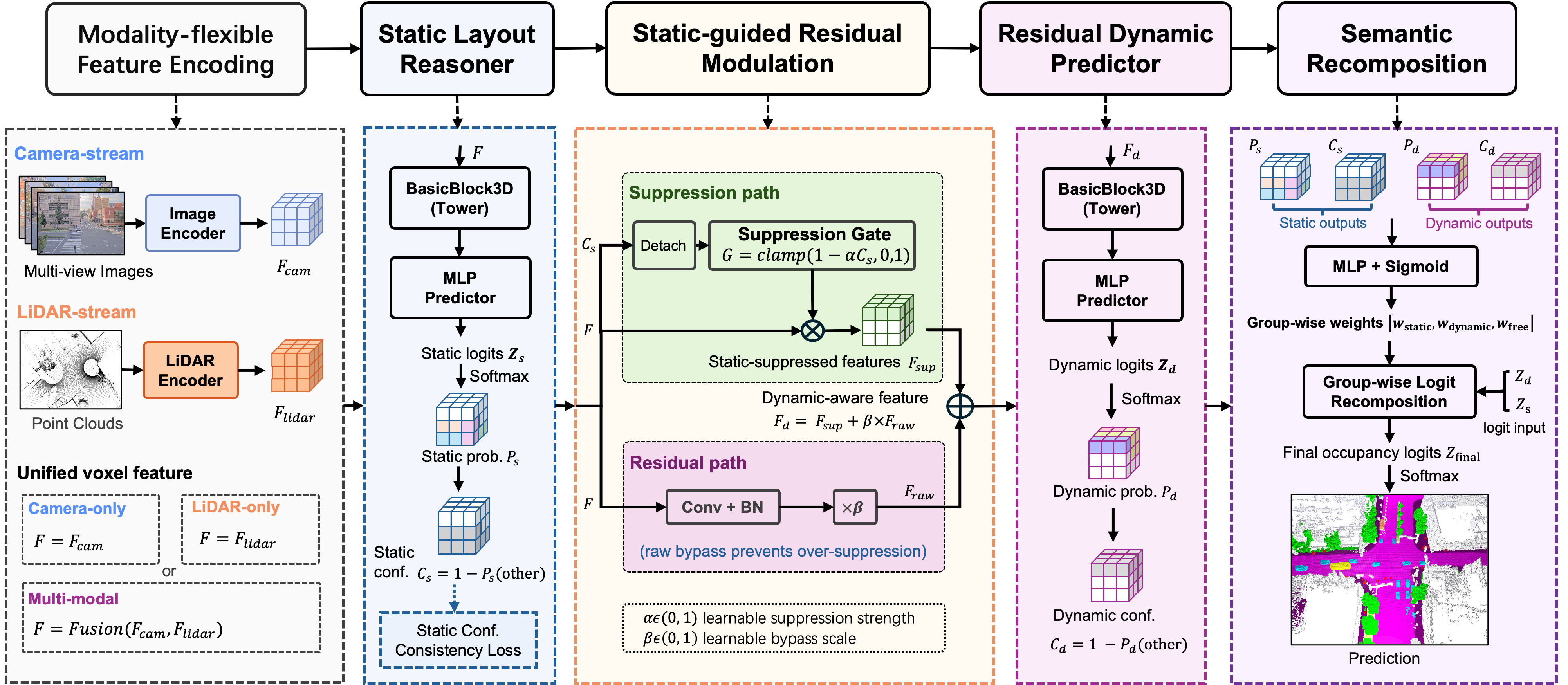}
\caption{\textbf{Overall framework of \nameplus{}.}
\nameplus{} formulates infrastructure-side occupancy prediction as a
progressive static-to-dynamic reasoning process rather than one-step voxel
classification. Given camera-only, LiDAR-only, or multi-modal inputs,
Modality-flexible Feature Encoding first produces a unified voxel representation
$\bm{F}$. Static Layout Reasoner then estimates static logits and confidence
$(\bm{Z}_{s}, \bm{C}_{s})$, which provide a soft explanation of persistent
infrastructure layout. Guided by this confidence, Static-guided Residual
Modulation suppresses static-dominant responses and derives dynamic-aware
residual features $\bm{F}_{d}$. Residual Dynamic Predictor further
recovers local dynamic evidence $(\bm{Z}_{d}, \bm{C}_{d})$, and Semantic
Recomposition integrates static, dynamic, and free-space evidence into
the final semantic occupancy logits $\bm{Z}_{\mathrm{final}}$, \rev{which are
converted by softmax into the occupancy prediction.}
}
\label{fig:method_overview_refined}
\end{figure*}

%% file: latex/tab/main-results.tex
\begin{table*}[!t]
\caption{\textbf{Main InfraOcc occupancy results.} \rev{Rankings are computed within each sensing track; the best, second-best, and third-best results are shaded red (bold), orange (underlined), and yellow (italic), respectively.}}
\label{tab:infraocc_main_results}
\centering
\definecolor{best}{RGB}{255, 180, 180}
\definecolor{second}{RGB}{255, 230, 180}
\definecolor{third}{RGB}{255, 255, 200}
\newlength{\mainmethodwd}
\newlength{\mainmetricwd}
\newlength{\mainclasswd}
\setlength{\mainmethodwd}{0.135\textwidth}
\setlength{\mainmetricwd}{0.061\textwidth}
\setlength{\mainclasswd}{\dimexpr(\textwidth-\mainmethodwd-4\mainmetricwd-2\arrayrulewidth)/15\relax}
\newcommand{\methodcell}[1]{\makebox[\mainmethodwd][l]{#1}}
\newcommand{\metriccell}[1]{\makebox[\mainmetricwd][c]{#1}}
\newcommand{\classcell}[1]{\makebox[\mainclasswd][c]{#1}}
\newcommand{\bestmetriccell}[1]{\cellcolor{best}\metriccell{\textbf{#1}}}
\newcommand{\secondmetriccell}[1]{\cellcolor{second}\metriccell{\underline{#1}}}
\newcommand{\thirdmetriccell}[1]{\cellcolor{third}\metriccell{\emph{#1}}}
\newcommand{\bestclasscell}[1]{\cellcolor{best}\classcell{\textbf{#1}}}
\newcommand{\secondclasscell}[1]{\cellcolor{second}\classcell{\underline{#1}}}
\newcommand{\thirdclasscell}[1]{\cellcolor{third}\classcell{\emph{#1}}}
\scriptsize
\setlength{\tabcolsep}{0pt}
\renewcommand{\arraystretch}{1.05}
\makebox[\textwidth][c]{%
\begin{tabular}{@{}c|cccc|*{15}{c}@{}}
\toprule
\methodcell{Method} & \metriccell{gIoU} & \metriccell{$\mathrm{mIoU}_{\mathrm{all}}$} & \metriccell{$\mathrm{mIoU}_{\mathrm{dyn}}$} & \metriccell{$\mathrm{mIoU}_{\mathrm{sta}}$} & \classcell{\occclassname{Bike}} & \classcell{\occclassname{Bus}} & \classcell{\occclassname{Car}} & \classcell{\occclassname{Motorcycle}} & \classcell{\occclassname{Pedestrian}} & \classcell{\occclassname{Truck}} & \classcell{\occclassname{Other}} & \classcell{\occclassname{Barrier}} & \classcell{\occclassname{Cone}} & \classcell{\occclassname{Driveable}} & \classcell{\occclassname{Sidewalk}} & \classcell{\occclassname{Terrain}} & \classcell{\occclassname{Manmade}} & \classcell{\occclassname{Vegetation}} & \classcell{\occclassname{Free}} \\
\methodcell{} & \metriccell{} & \metriccell{} & \metriccell{} & \metriccell{} & \classcell{\classswatch{occBike}} & \classcell{\classswatch{occBus}} & \classcell{\classswatch{occCar}} & \classcell{\classswatch{occMotor}} & \classcell{\classswatch{occPed}} & \classcell{\classswatch{occTruck}} & \classcell{\classswatch{occOther}} & \classcell{\classswatch{occBarrier}} & \classcell{\classswatch{occCone}} & \classcell{\classswatch{occDrive}} & \classcell{\classswatch{occSide}} & \classcell{\classswatch{occTerrain}} & \classcell{\classswatch{occManmade}} & \classcell{\classswatch{occVegetation}} & \classcell{\classswatch{occFree}} \\
\midrule
\rowcolor{gray!5}{\methodcell{\textit{Multi-modal (C+L)}}}  \\
\methodcell{BEVFusion~\cite{BEVFusion}} & \metriccell{\secondcell{77.01}} & \metriccell{\secondcell{55.44}} & \metriccell{\secondcell{32.57}} & \metriccell{\secondcell{72.60}} & \classcell{\secondcell{11.03}} & \classcell{\secondcell{65.04}} & \classcell{\secondcell{55.73}} & \classcell{\secondcell{15.26}} & \classcell{\secondcell{29.48}} & \classcell{\bestcell{18.88}} & \classcell{\secondcell{55.04}} & \classcell{\secondcell{72.04}} & \classcell{\secondcell{72.07}} & \classcell{\secondcell{75.43}} & \classcell{\secondcell{73.53}} & \classcell{\secondcell{81.26}} & \classcell{\secondcell{76.82}} & \classcell{\secondcell{74.60}} & \classcell{\secondcell{96.58}} \\
\rowcolor{gray!5}\methodcell{ProSD-Occ} & \metriccell{\bestcell{91.78}} & \metriccell{\bestcell{65.87}} & \metriccell{\bestcell{37.92}} & \metriccell{\bestcell{86.82}} & \classcell{\bestcell{30.07}} & \classcell{\bestcell{66.95}} & \classcell{\bestcell{58.55}} & \classcell{\bestcell{19.82}} & \classcell{\bestcell{34.65}} & \classcell{\secondcell{17.51}} & \classcell{\bestcell{72.31}} & \classcell{\bestcell{73.79}} & \classcell{\bestcell{89.68}} & \classcell{\bestcell{94.25}} & \classcell{\bestcell{88.35}} & \classcell{\bestcell{94.73}} & \classcell{\bestcell{90.37}} & \classcell{\bestcell{91.12}} & \classcell{\bestcell{99.09}} \\
\midrule
\rowcolor{gray!5}{\methodcell{\textit{LiDAR-Only (L)}}}  \\
\methodcell{VoxelNet~\cite{VoxelNet}} & \metriccell{\secondcell{75.69}} & \metriccell{\secondcell{53.87}} & \metriccell{31.13} & \metriccell{\secondcell{70.92}} & \classcell{11.08} & \classcell{\bestcell{65.63}} & \classcell{\secondcell{54.52}} & \classcell{\secondcell{13.43}} & \classcell{27.85} & \classcell{14.29} & \classcell{\secondcell{53.39}} & \classcell{72.22} & \classcell{\secondcell{68.26}} & \classcell{\secondcell{75.13}} & \classcell{\secondcell{72.35}} & \classcell{\secondcell{78.99}} & \classcell{\secondcell{74.69}} & \classcell{\secondcell{72.36}} & \classcell{\secondcell{96.31}} \\
\methodcell{PointPillars~\cite{PointPillars}} & \metriccell{70.53} & \metriccell{50.51} & \metriccell{\secondcell{31.55}} & \metriccell{64.74} & \classcell{\bestcell{14.99}} & \classcell{61.62} & \classcell{54.03} & \classcell{11.50} & \classcell{\secondcell{32.62}} & \classcell{\secondcell{14.55}} & \classcell{44.06} & \classcell{\bestcell{73.83}} & \classcell{57.49} & \classcell{72.11} & \classcell{67.67} & \classcell{70.75} & \classcell{66.42} & \classcell{65.56} & \classcell{95.27} \\
\rowcolor{gray!5}\methodcell{ProSD-Occ} & \metriccell{\bestcell{90.86}} & \metriccell{\bestcell{63.66}} & \metriccell{\bestcell{33.87}} & \metriccell{\bestcell{85.99}} & \classcell{\secondcell{11.59}} & \classcell{\secondcell{65.62}} & \classcell{\bestcell{58.29}} & \classcell{\bestcell{15.22}} & \classcell{\bestcell{33.09}} & \classcell{\bestcell{19.42}} & \classcell{\bestcell{68.00}} & \classcell{\secondcell{73.75}} & \classcell{\bestcell{90.18}} & \classcell{\bestcell{94.02}} & \classcell{\bestcell{88.14}} & \classcell{\bestcell{95.16}} & \classcell{\bestcell{88.53}} & \classcell{\bestcell{90.17}} & \classcell{\bestcell{98.99}} \\
\midrule
\rowcolor{gray!5}{\methodcell{\textit{Camera-Only (C)}}} \\
\methodcell{BEVDet~\cite{BEVDet}} & \metriccell{68.62} & \metriccell{45.14} & \thirdmetriccell{21.19} & \metriccell{63.11} & \classcell{9.23} & \thirdclasscell{42.75} & \classcell{38.04} & \thirdclasscell{9.68} & \classcell{15.01} & \secondclasscell{12.45} & \classcell{42.88} & \classcell{70.36} & \classcell{53.77} & \classcell{72.04} & \classcell{68.07} & \classcell{73.09} & \classcell{64.28} & \classcell{60.38} & \classcell{94.87} \\
\methodcell{BEVFormer~\cite{BEVFormer}} & \thirdmetriccell{81.72} & \metriccell{47.76} & \metriccell{10.73} & \thirdmetriccell{75.53} & \classcell{7.97} & \classcell{20.39} & \classcell{17.65} & \classcell{3.47} & \classcell{8.62} & \classcell{6.30} & \classcell{42.47} & \thirdclasscell{76.26} & \secondclasscell{72.83} & \classcell{85.19} & \thirdclasscell{81.95} & \thirdclasscell{88.95} & \thirdclasscell{79.52} & \thirdclasscell{77.05} & \thirdclasscell{97.56} \\
\methodcell{BEVDepth~\cite{BEVDepth}} & \metriccell{71.89} & \metriccell{46.81} & \metriccell{19.64} & \metriccell{67.19} & \secondclasscell{14.84} & \classcell{33.87} & \classcell{38.74} & \classcell{3.42} & \classcell{16.86} & \classcell{10.14} & \classcell{48.73} & \classcell{71.05} & \classcell{62.37} & \classcell{73.31} & \classcell{69.32} & \classcell{76.07} & \classcell{70.51} & \classcell{66.15} & \classcell{95.54} \\
\methodcell{TPVFormer~\cite{TPVFormer}} & \metriccell{73.80} & \metriccell{43.09} & \metriccell{4.36} & \metriccell{72.13} & \classcell{0.10} & \classcell{3.69} & \classcell{13.80} & \classcell{0.41} & \classcell{6.86} & \classcell{1.33} & \classcell{42.86} & \classcell{75.29} & \classcell{69.69} & \classcell{80.61} & \classcell{77.62} & \classcell{86.63} & \classcell{72.61} & \classcell{71.70} & \classcell{96.12} \\
\methodcell{SurroundOcc~\cite{SurroundOcc}} & \secondmetriccell{83.78} & \secondmetriccell{50.77} & \metriccell{11.37} & \secondmetriccell{80.32} & \classcell{12.66} & \classcell{21.90} & \classcell{18.54} & \classcell{2.06} & \classcell{9.50} & \classcell{3.55} & \thirdclasscell{65.09} & \bestclasscell{79.04} & \thirdclasscell{72.70} & \thirdclasscell{86.38} & \secondclasscell{83.99} & \secondclasscell{91.10} & \secondclasscell{83.16} & \secondclasscell{81.09} & \secondclasscell{97.80} \\
\methodcell{CONet~\cite{OpenOccupancy}} & \metriccell{78.56} & \metriccell{49.25} & \metriccell{18.93} & \metriccell{72.00} & \classcell{3.06} & \classcell{37.24} & \thirdclasscell{39.47} & \classcell{4.42} & \thirdclasscell{17.64} & \thirdclasscell{11.72} & \classcell{45.01} & \classcell{75.31} & \classcell{64.94} & \classcell{84.29} & \classcell{79.45} & \classcell{84.27} & \classcell{72.96} & \classcell{69.79} & \classcell{97.08} \\
\methodcell{STCOcc~\cite{STCOcc}} & \metriccell{73.11} & \thirdmetriccell{49.70} & \secondmetriccell{23.45} & \metriccell{69.39} & \thirdclasscell{13.33} & \secondclasscell{49.57} & \secondclasscell{42.64} & \secondclasscell{10.37} & \secondclasscell{18.09} & \classcell{6.67} & \classcell{52.63} & \classcell{71.94} & \classcell{67.09} & \secondclasscell{86.83} & \classcell{74.94} & \classcell{80.54} & \classcell{57.27} & \classcell{63.91} & \classcell{96.89} \\
\methodcell{SparseOcc~\cite{SparseOcc}} & \metriccell{71.61} & \metriccell{45.42} & \metriccell{9.41} & \metriccell{72.42} & \classcell{1.50} & \classcell{20.94} & \classcell{24.87} & \classcell{0.00} & \classcell{4.60} & \classcell{4.57} & \bestclasscell{67.68} & \secondclasscell{76.37} & \classcell{61.51} & \classcell{74.67} & \classcell{76.59} & \classcell{88.54} & \classcell{68.79} & \classcell{65.19} & \classcell{96.31} \\
\methodcell{GaussianFormer~\cite{GaussianFormer}} & \metriccell{77.89} & \metriccell{43.63} & \metriccell{10.45} & \metriccell{68.51} & \classcell{3.13} & \classcell{22.11} & \classcell{23.17} & \classcell{1.29} & \classcell{8.21} & \classcell{4.80} & \classcell{36.70} & \classcell{68.54} & \classcell{55.60} & \classcell{82.59} & \classcell{79.37} & \classcell{74.20} & \classcell{77.85} & \classcell{73.21} & \classcell{96.71} \\
\rowcolor{gray!5}\methodcell{ProSD-Occ} & \bestmetriccell{89.08} & \bestmetriccell{60.08} & \bestmetriccell{28.97} & \bestmetriccell{83.42} & \bestclasscell{22.74} & \bestclasscell{55.45} & \bestclasscell{47.84} & \bestclasscell{12.27} & \bestclasscell{21.01} & \bestclasscell{14.49} & \secondclasscell{65.99} & \classcell{72.72} & \bestclasscell{82.14} & \bestclasscell{93.84} & \bestclasscell{86.88} & \bestclasscell{93.31} & \bestclasscell{86.09} & \bestclasscell{86.38} & \bestclasscell{98.78} \\
\bottomrule
\end{tabular}%
}
\end{table*}

%% file: latex/4-exps.tex
\subsection{Experimental Setup}
\noindent \rev{\textbf{Protocol.}
We evaluate camera-only (C), LiDAR-only (L), and multi-modal (C+L) occupancy
on the InfraOcc test split using the unified protocol in
Sec.~\ref{sec:benchmark_eval}. Baselines are retrained using their official
implementations under the same budget, and the suffixes (C), (L), and (C+L)
identify the sensing track. Controlled ablations use the camera-only setting,
where weak depth evidence makes separating transient objects from persistent
layout particularly difficult.
\emph{Plain (C)} is a flat one-stage counterpart that removes the proposed
progressive static-to-dynamic reasoning mechanism and directly predicts all
occupancy classes from the same encoded voxel representation; the feature
encoder and final occupancy loss remain unchanged.}\\
\noindent \rev{\textbf{Implementation Details.} The camera branch uses
an ImageNet-pretrained ResNet-50 at a $256\times704$ input resolution with
depth bins of 1--60\,m at 1\,m intervals; the LiDAR branch voxelizes point
clouds at $0.1$\,m. The static and dynamic towers use 40 and 64 hidden
channels, and the suppression gate is initialized at $\alpha{=}0.5$. All
supervision terms use unit weights ($\lambda_s{=}\lambda_d{=}1$) with
frequency-based class weights, and the static-consistency regularizer uses
$\lambda_c{=}0.1$ and $\tau{=}0.7$. Models are trained for 32 epochs with
AdamW~\cite{AdamW} (learning rate $1\times10^{-4}$, weight decay $10^{-2}$),
a cosine-annealing schedule, and random image and BEV flip augmentation, on
8 NVIDIA A100 GPUs with a per-GPU batch size of 2.}\\
\noindent \rev{\textbf{Metrics.}
Following Sec.~\ref{sec:benchmark_eval}, we report gIoU,
$\mathrm{mIoU}_{\mathrm{all}}$ (All), $\mathrm{mIoU}_{\mathrm{dyn}}$ (Dyn),
and $\mathrm{mIoU}_{\mathrm{sta}}$ (Sta), with per-class and free-space IoU in
the main comparison. Dyn is the primary diagnostic for sparse traffic
participants; Sta checks static-layout preservation, whereas gIoU and
free-space IoU measure geometric occupancy.}

\input{latex/fig/qualitative-comparison}

\input{latex/tab/core-mechanism-ablation}
\subsection{Main Results}
\label{sec:exp_quantitative}
\subsubsection{Quantitative Comparison}

\rev{As reported in Tab.~\ref{tab:infraocc_main_results}, in the camera-only
track, where static background dominance is most severe, \nameplus{} reaches
60.08 All and 28.97 Dyn, surpassing the best previous camera results by
+9.31 All (over SurroundOcc~\cite{SurroundOcc}) and +5.52 Dyn (over
STCOcc~\cite{STCOcc}), while also achieving the best Sta (83.42) and gIoU
(89.08).} The simultaneous gains in
Dyn, Sta, and gIoU are important for the paper's thesis: \nameplus{} does not
obtain better mIoU by simply fitting the persistent background, and it does not
recover dynamic objects by sacrificing layout quality.
The LiDAR-only and multi-modal tracks show that the progressive reasoning
framework remains effective with stronger geometric evidence. \nameplus{} reaches
63.66/33.87 All/Dyn in the L setting and 65.87/37.92 in the C+L setting,
outperforming the corresponding listed baselines in both overall and dynamic
occupancy. The lower camera-only Dyn score reflects the difficulty of
monocular roadside depth inference, while the consistent ranking across C, L,
and C+L indicates that the contribution is not tied to one sensor front-end.
Instead, the gain comes from organizing voxel evidence from persistent static
layout toward residual dynamic occupancy.

\subsubsection{Qualitative Comparison}

\rev{The qualitative comparison in \Cref{fig:infraocc_modality_qualitative}
aligns roadside camera
evidence, ground-truth occupancy, and four camera-only predictions on the
same test samples, making the static-dynamic asymmetry directly visible.
TPVFormer~\cite{TPVFormer} and SparseOcc~\cite{SparseOcc} reconstruct the
persistent road layout reasonably
well but almost entirely miss the vehicles and pedestrians crossing the
intersection, mirroring their low dynamic mIoU in
Tab.~\ref{tab:infraocc_main_results}. STCOcc~\cite{STCOcc} recovers more dynamic
occupancy, but nearby traffic participants are often blended into fragmented
blobs with confused semantic labels, leaving different dynamic categories
indistinguishable along lanes and crosswalks. In contrast, \nameplus{} (C)
produces well-separated, category-consistent dynamic occupancy---vehicle
clusters and crossing pedestrians remain individually
identifiable---while keeping sidewalk and vegetation boundaries closest to
the ground truth. This indicates that the dynamic gain comes from exposing
residual dynamic evidence rather than from trading away static structure.}

\input{latex/fig/mechanism-feature-visualization}

\subsection{Ablation Studies}
\label{sec:exp_ablation}
\subsubsection{Progressive Static-to-Dynamic Reasoning}
\rev{The central causal claim is that dynamic occupancy improves when the
model reasons through static layout before predicting residual objects,
which we test in Tab.~\ref{tab:core_mechanism}.} \rev{The Plain (C) baseline
achieves 50.71 All and 23.48
Dyn. The parallel static-to-dynamic (S2D) fusion control reaches only 53.59
All and 24.43 Dyn;}
this limited gain shows that independent \rev{static-dynamic} prediction is not
sufficient without residual conditioning.
Progressive reasoning further reaches 60.08 All and 28.97 Dyn because the
dynamic branch no longer competes with the full static scene in one step; it
receives a layout-conditioned feature where residual object evidence is easier
to recover.

The lower block explains why this residual feature must preserve both
suppressed and raw evidence. Static-guided suppression exposes dynamic cues and
improves Dyn from 24.43 to 28.81, but suppression alone weakens the information
needed for complete semantic reconstruction. Adding the raw-feature bypass
produces the full result, raising Sta to 83.42 while keeping the best Dyn.
This supports the intended role of the module: static evidence is not removed;
it is attenuated where it dominates and then recombined with residual evidence.
\Cref{fig:prosd_mechanism_features} visualizes the same mechanism across
multiple samples. Static confidence is high on persistent road and
infrastructure regions, suppression is applied around static-dominant
responses, and the Dynamic-aware feature, visualized as weighted branch-feature
energy, remains spatially continuous while emphasizing sparse foreground
occupancy. The last column shows the corresponding ground-truth dynamic
foreground to make the sparse target regions explicit.
\input{latex/fig/mechanism-error-comparison}

\rev{A direct error comparison between Plain (C)
and \nameplus{} (C) is given in \Cref{fig:prosd_mechanism_errors}.} The error maps show that residual
modulation reduces missed and false dynamic occupancy without turning the static
branch into a memorized background mask.

\subsubsection{Guidance, Suppression, and Recomposition}
\rev{We next decompose the residual recovery path into guidance, attenuation,
and recomposition in
Tabs.~\ref{tab:static_guidance_quality}--\ref{tab:adaptive_fusion_design}.
As shown in Tab.~\ref{tab:static_guidance_quality}, static guidance is
necessary, and its quality matters.} With
guidance disabled, the row uses the shared no-static-suppression control
reported in Tabs.~\ref{tab:core_mechanism} and~\ref{tab:suppression_strength},
giving 53.59 All and 24.43 Dyn. Noisy guidance further harms dynamic
prediction, and learned static confidence reaches 28.97 Dyn, \rev{only 0.21
below oracle guidance, indicating that the learned confidence has nearly
saturated the attainable guidance quality and the remaining headroom lies in
residual reasoning rather than in better guidance}. Thus the static
branch is not merely an auxiliary prediction pathway; it provides a sample-adaptive
soft explanation of persistent layout that makes residual dynamic evidence
easier to locate.

\rev{As shown in Tab.~\ref{tab:suppression_strength}, the attenuation itself
should be learned rather than fixed.} Fixed suppression improves Dyn over no suppression,
from 24.43 to 28.32, confirming that static confidence is useful as a soft mask
for exposing residual foreground cues. However, a fixed gate cannot adapt to
scene-dependent ambiguity and remains limited in All/Sta. In contrast,
learnable suppression raises All/Sta to 60.08/83.42 while also giving the best
Dyn among practical settings. The initialization study reinforces the same
interpretation: both weaker ($\alpha_{\mathrm{init}}=0.25$) and stronger
($\alpha_{\mathrm{init}}=0.75$) starting points preserve reasonable Dyn, but
they reduce All and Sta. This indicates that the gate must balance residual
mining and layout preservation, rather than simply maximizing foreground
response or suppressing static features aggressively.

\rev{The final recomposition step is verified in
Tab.~\ref{tab:adaptive_fusion_design}.} The differences are moderate rather than dominant:
deterministic average and deterministic group-wise merging trail the default by
1.79 and 1.33 All, respectively, showing that fixed recomposition is
suboptimal but not catastrophic. One-input gates also lose about 1.1--1.5 All,
while class-wise routing remains closer to the default but still weakens
All/Sta. This pattern is consistent with adaptive fusion being an auxiliary
recomposition design: it stabilizes the final static/dynamic routing, but the
main improvement still comes from progressive residual reasoning.

\input{latex/tab/static-guidance-quality}
\input{latex/tab/suppression-strength}
\input{latex/tab/adaptive-fusion-design}
\input{latex/tab/static-consistency}

\subsubsection{Static Consistency}
\rev{We further check in Tab.~\ref{tab:static_consistency} whether the
static scaffold used for residual modulation should be regularized across
samples.} Removing consistency
reduces All from 60.08 to 58.34 and Sta from 83.42 to 80.95, and also lowers
Dyn from 28.97 to 28.20. \rev{In the design block, batch-level consensus is
the default consistency form (Sec.~\ref{sec:prosd_objective}) and
matches the static-consistency row. Pairwise agreement remains close but is
slightly weaker at 59.81 All, 28.74 Dyn, and 83.12 Sta: a single partner
frame is a noisier target under transient occlusions and sensing noise,
whereas the batch consensus averages such fluctuations into a
more stable target. The final model therefore keeps batch-level consensus as
the default static regularizer.}

\input{latex/tab/loss-balance}
\input{latex/tab/loss-components}
\input{latex/tab/accuracy-efficiency}
\input{latex/tab/distance-wise-dynamic}

\subsubsection{Loss Design}
\rev{A natural concern is whether the dynamic gain could be obtained by loss
reweighting alone, which we examine in Tab.~\ref{tab:loss_balance}.} Dynamic-heavy training gives the highest Dyn
(29.26), but its All and Sta are lower than the balanced setting. Conversely,
dynamic-light or static-heavy settings preserve more layout structure but
weaken dynamic recovery. The balanced objective is therefore used because it
best supports the coupled roles of the two branches: static layout explains the
persistent scaffold, while dynamic supervision recovers sparse residual
objects.

\rev{The loss composition for each supervision target is studied in
Tab.~\ref{tab:loss_terms}.} CE alone provides category supervision but is insufficient under the
severe static-dynamic imbalance of roadside scenes. Semantic scaling improves
calibration under class imbalance, and Lovasz improves region-level overlap for
fragmented dynamic objects. The full CE+Sem.+Lovasz setting gives the strongest
branch results, while the fusion-output block confirms that the final
recomposed occupancy field also needs structured supervision rather than a
single voxel-wise classification loss.

\subsubsection{Efficiency}

\rev{As shown in Tab.~\ref{tab:capacity_efficiency}, the gain is not a
consequence of indiscriminate model scaling.} In the camera-only setting, \rev{\nameplus{} (C)} uses
fewer parameters than BEVDet, BEVDepth, SparseOcc, and CONet, while achieving
the best All and Dyn among the listed methods. In the LiDAR-only and
multi-modal settings, the method also remains competitive in parameter count
while giving substantially stronger occupancy accuracy. These results match the
contribution of the paper: \nameplus{} changes how voxel evidence is organized
at the decision stage, rather than relying on a larger feature extractor.

\subsubsection{Distance-wise Dynamic Occupancy}

\rev{The distance-binned evaluation in Tab.~\ref{tab:distance_robustness}
locates where dynamic occupancy remains difficult after the main gains.} \rev{\nameplus{} (C)} obtains 29.66 Dyn in the 0--20\,m
range and 31.90 in the 20--40\,m range, but drops to 8.24 in the 40--60\,m
range. Adding explicit geometry improves the far-range bin substantially:
\rev{\nameplus{} (L)} reaches 35.36/34.08/21.22 over the three distance bins, and \rev{\nameplus{} (C+L)}
further reaches 37.58/39.60/23.28. This trend clarifies the scope of the
method. Progressive static-to-dynamic reasoning improves semantic organization
across sensing regimes, but long-range dynamic occupancy remains limited by the
available geometric evidence.

\input{latex/tab/background-overfitting-robustness}
\input{latex/fig/robustness-curves}
\subsubsection{Robustness to Static-Background Overfitting}
\rev{A risk specific to fixed-view infrastructure perception is that a model
may appear strong simply by associating fixed coordinates with persistent
background, which we probe in Tab.~\ref{tab:background_overfitting}.} We apply controlled rigid
\rev{coordinate-frame} perturbations at test time by translating or yaw-rotating the
evaluation frame and applying the same re-anchoring to the sensor pose metadata
and labels. The physical scene is unchanged, but direct coordinate-background
shortcuts become less reliable. We therefore report dynamic mIoU after
re-anchoring: if dynamic recovery is driven by sensor evidence rather than
static-coordinate memorization, it should remain competitive when the static
scaffold is no longer aligned with its original coordinates.
\rev{The clean column reproduces the unperturbed results
(Tab.~\ref{tab:infraocc_main_results} for \nameplus{} and
Tab.~\ref{tab:core_mechanism} for Plain (C)). In the camera-only
setting, \nameplus{} (C) starts from a stronger clean dynamic score and keeps
higher dynamic prediction than Plain (C) under all translation and rotation
perturbations. The \nameplus{} (L) and (C+L) variants start from stronger
clean scores and remain ahead of the camera model in most re-anchored
settings, but degrade faster at the largest shifts, indicating that explicit
geometry strengthens dynamic evidence yet is more sensitive to
coordinate-frame misalignment.}
\Cref{fig:robustness_c_prosd_vs_plain} makes this trade-off explicit by
plotting \nameplus{} (C) and Plain (C) as paired absolute dynamic-mIoU curves.
Across all tested translation and rotation shifts, the dynamic gap remains
positive, supporting the intended interpretation of ProSD reasoning: it improves
residual dynamic recovery from sensor evidence under controlled coordinate
re-anchoring.

%% file: latex/fig/qualitative-comparison.tex
\begin{figure*}[!t]
  \centering
  \includegraphics[width=0.98\textwidth]{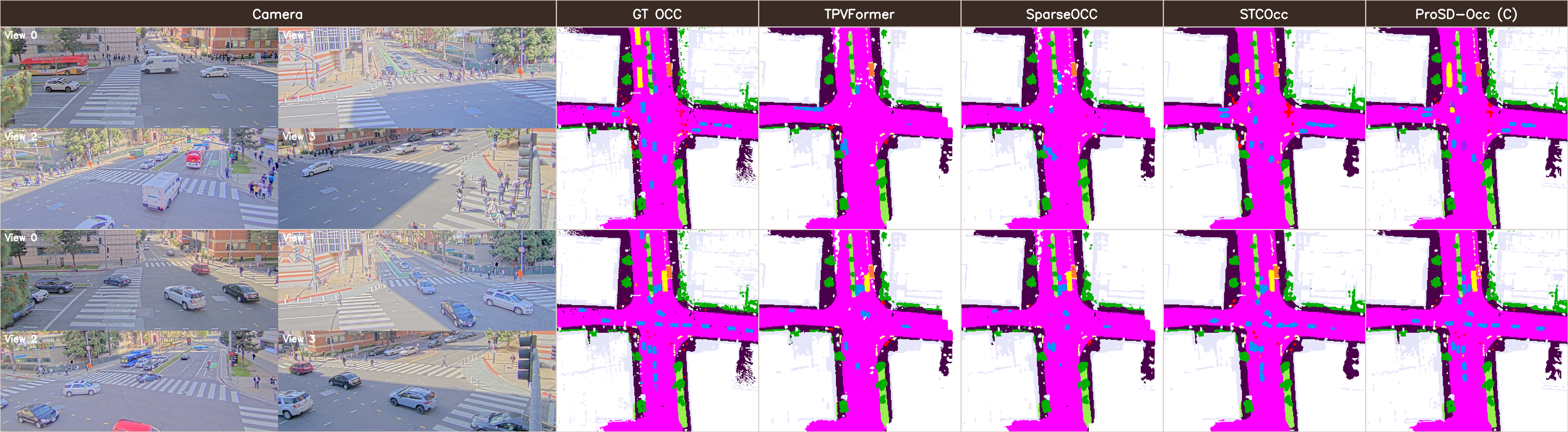}
  \caption{\textbf{\rev{Qualitative comparison of camera-only methods.}}
  \rev{Each row shows a test sample} with aligned roadside camera evidence,
  ground-truth occupancy, and camera-only occupancy predictions from
  TPVFormer~\cite{TPVFormer},
  \rev{SparseOcc}~\cite{SparseOcc}, STCOcc~\cite{STCOcc}, and ProSD-Occ (C).}
  \label{fig:infraocc_modality_qualitative}
\end{figure*}

%% file: latex/tab/core-mechanism-ablation.tex
\begin{table}[!tbp]
\caption{\textbf{Core ablation of progressive static-to-dynamic (S2D) reasoning.} The upper block compares prediction paradigms; the lower block isolates suppression and raw-feature bypass. The parallel S2D control uses the same no-static-suppression setting as the lower-block control.}
\label{tab:core_mechanism}
\centering
\scriptsize
\setlength{\tabcolsep}{4.2pt}
\renewcommand{\arraystretch}{1.08}
\begin{tabularx}{\columnwidth}{@{}>{\raggedright\arraybackslash}Xcccc@{}}
\toprule
Setting & gIoU$\uparrow$ & $\mathrm{mIoU}_{\mathrm{all}}\uparrow$ & $\mathrm{mIoU}_{\mathrm{dyn}}\uparrow$ & $\mathrm{mIoU}_{\mathrm{sta}}\uparrow$ \\
\midrule
\multicolumn{5}{l}{\textbf{\textit{Necessity of Progressive S2D Reasoning}}} \\
\rev{Plain (C)} & 78.54 & 50.71 & 23.48 & 71.14 \\
Parallel S2D Fusion (no supp.) & 79.77 & 53.59 & 24.43 & 75.46 \\
Progressive S2D Reasoning & 89.08 & 60.08 & 28.97 & 83.42 \\
\midrule
\multicolumn{5}{l}{\textbf{\textit{Design of Progressive S2D Reasoning}}} \\
No static suppression & 79.77 & 53.59 & 24.43 & 75.46 \\
Suppression only & 88.35 & 59.77 & 28.81 & 82.99 \\
Suppression + raw bypass & 89.08 & 60.08 & 28.97 & 83.42 \\
\bottomrule
\end{tabularx}
\end{table}

%% file: latex/fig/mechanism-feature-visualization.tex
\begin{figure}[!t]
  \centering
  \includegraphics[width=\columnwidth]{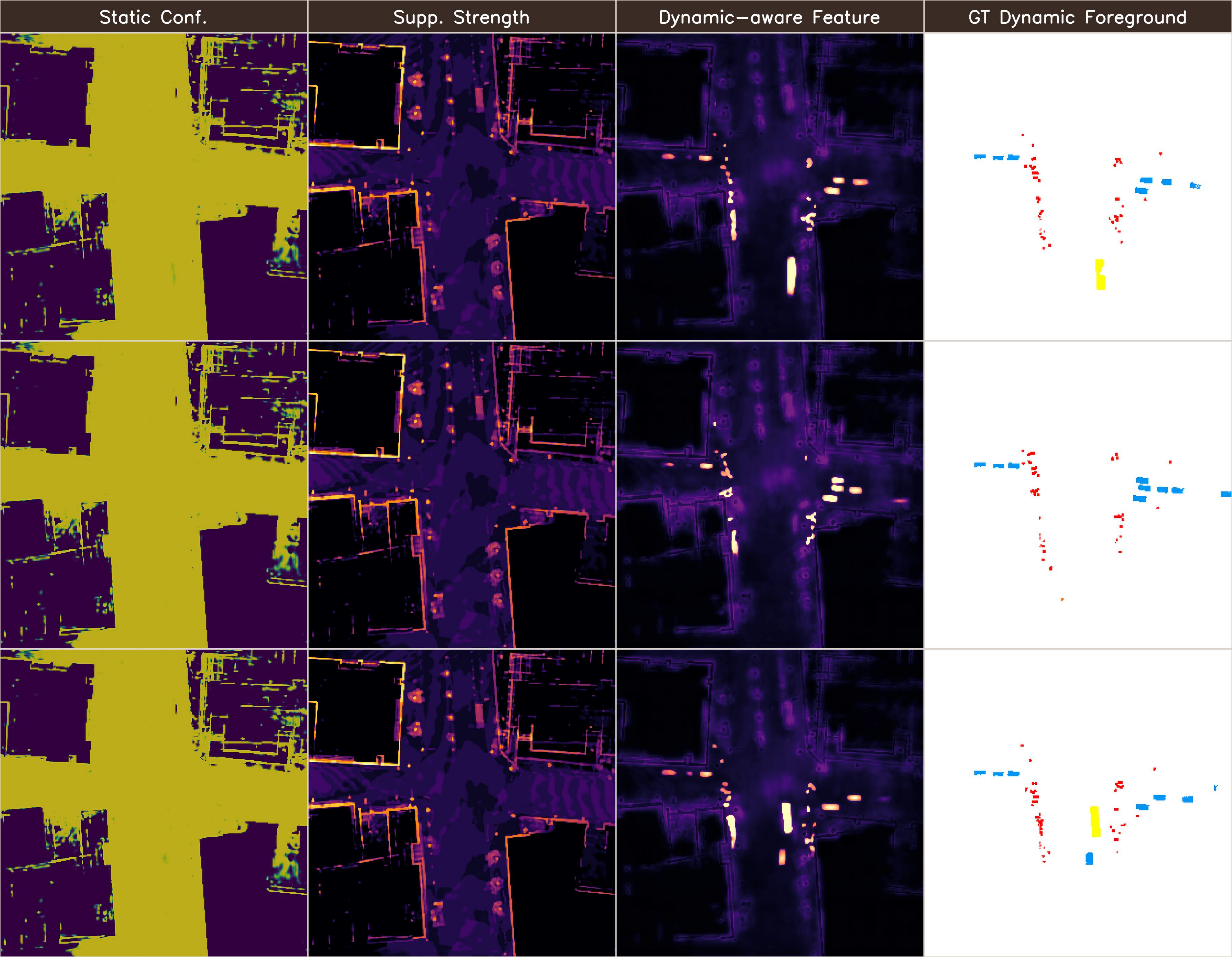}
  \caption{\textbf{Progressive static-to-dynamic feature visualization.}
  Each row shows one \rev{test sample}. \rev{The columns visualize static
  confidence, static-confidence suppression strength, the dynamic-aware
  feature energy after static suppression, and the corresponding ground-truth
  dynamic foreground occupancy. Static confidence is high on persistent
  structures, and suppression exposes the sparse dynamic foreground.}}
  \label{fig:prosd_mechanism_features}
\end{figure}

%% file: latex/fig/mechanism-error-comparison.tex
\begin{figure}[!t]
  \centering
  \includegraphics[width=\columnwidth]{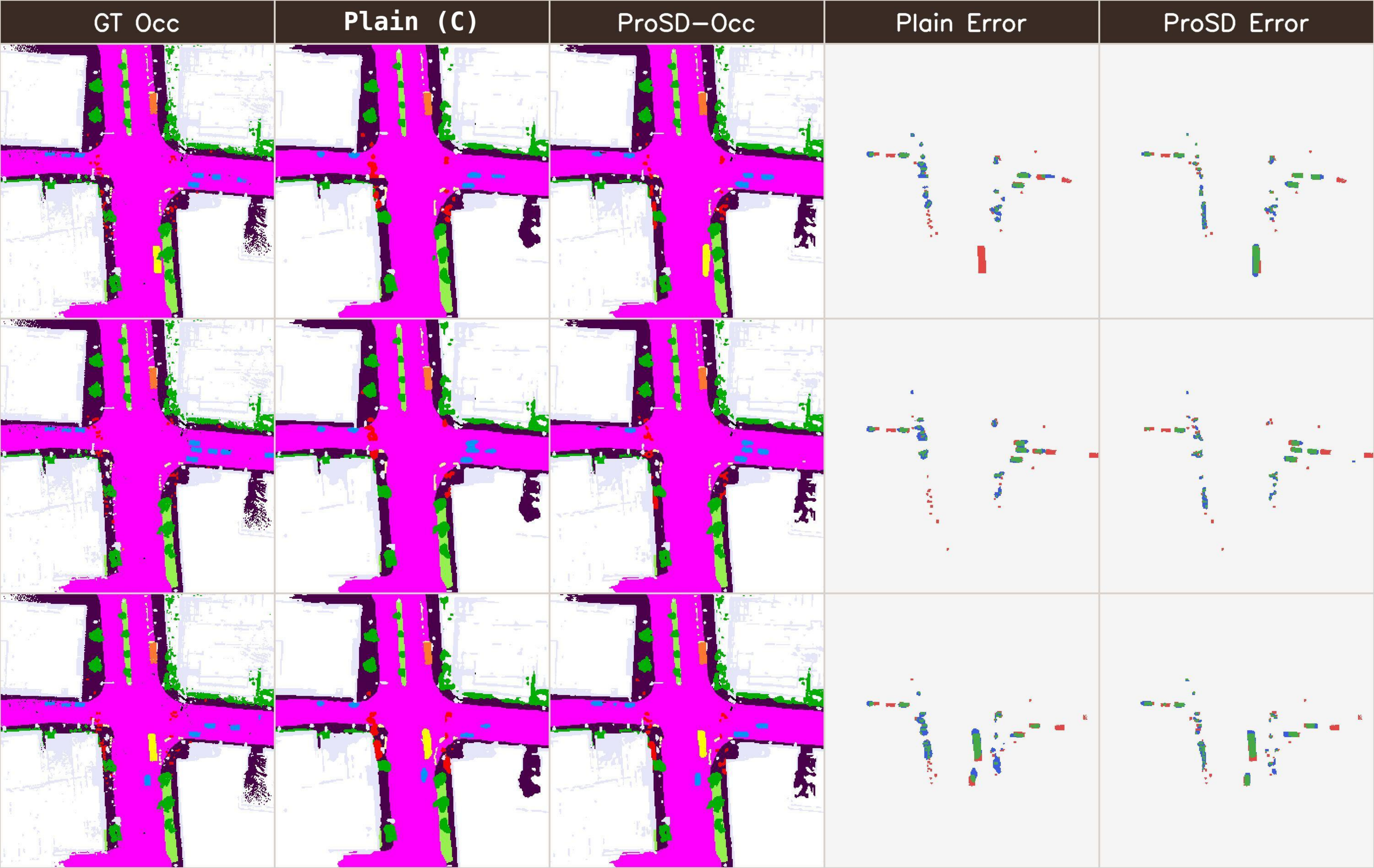}
  \caption{\textbf{Dynamic occupancy recovery by progressive reasoning.}
  Each row uses the same sample as \Cref{fig:prosd_mechanism_features},
  \rev{comparing Plain (C) with \nameplus{} (C)}.
  The error maps compare dynamic
  foreground occupancy only: green indicates correct dynamic prediction, blue
  indicates missed dynamic occupancy, and red indicates false dynamic
   prediction.}
  \label{fig:prosd_mechanism_errors}
\end{figure}

%% file: latex/tab/static-guidance-quality.tex
\begin{table}[!tbp]
\caption{\textbf{Static guidance quality.} Only the guidance signal for residual modulation is changed.
None uses the shared no-static-suppression control, Noisy uses a perturbed static-confidence map,
Predicted uses learned static confidence from \rev{the static branch}, and Oracle
uses ground-truth static occupancy as guidance.}
\label{tab:static_guidance_quality}
\centering
\scriptsize
\setlength{\tabcolsep}{2.8pt}
\renewcommand{\arraystretch}{1.08}
\begin{tabular*}{\columnwidth}{@{\extracolsep{\fill}}lcccc@{}}
\toprule
Guidance & gIoU$\uparrow$ & $\mathrm{mIoU}_{\mathrm{all}}\uparrow$ & $\mathrm{mIoU}_{\mathrm{dyn}}\uparrow$ & $\mathrm{mIoU}_{\mathrm{sta}}\uparrow$ \\
\midrule
None & 79.77 & 53.59 & 24.43 & 75.46 \\
Noisy & 86.70 & 56.51 & 22.60 & 81.95 \\
Predicted & 89.08 & 60.08 & 28.97 & 83.42 \\
Oracle & 89.22 & 60.31 & 29.18 & 83.65 \\
\bottomrule
\end{tabular*}
\end{table}

%% file: latex/tab/suppression-strength.tex
\begin{table}[!tbp]
\caption{\textbf{Suppression strength design.} We compare static-guided attenuation choices and initialization sensitivity.}
\label{tab:suppression_strength}
\centering
\scriptsize
\setlength{\tabcolsep}{2.3pt}
\renewcommand{\arraystretch}{1.08}
\begin{tabular*}{\columnwidth}{@{\extracolsep{\fill}}lccccc@{}}
\toprule
Setting & $\alpha_{\mathrm{init}}$ & gIoU$\uparrow$ & $\mathrm{mIoU}_{\mathrm{all}}\uparrow$ & $\mathrm{mIoU}_{\mathrm{dyn}}\uparrow$ & $\mathrm{mIoU}_{\mathrm{sta}}\uparrow$ \\
\midrule
\multicolumn{6}{l}{\textbf{\textit{Necessity of learnable suppression}}} \\
No static suppression & 0.0 & 79.77 & 53.59 & 24.43 & 75.46 \\
Fixed suppression & 0.5 & 84.30 & 56.68 & 28.32 & 77.95 \\
Learnable suppression & 0.5 & 89.08 & 60.08 & 28.97 & 83.42 \\
\midrule
\multicolumn{6}{l}{\textbf{\textit{Sensitivity to initialization}}} \\
Learnable ($\alpha_{\mathrm{init}}=0.25$) & 0.25 & 86.42 & 58.27 & 28.72 & 80.44 \\
Learnable ($\alpha_{\mathrm{init}}=0.50$) & 0.50 & 89.08 & 60.08 & 28.97 & 83.42 \\
Learnable ($\alpha_{\mathrm{init}}=0.75$) & 0.75 & 86.10 & 57.91 & 28.64 & 79.86 \\
\bottomrule
\end{tabular*}
\end{table}

%% file: latex/tab/adaptive-fusion-design.tex
\begin{table}[!tbp]
\caption{\textbf{Adaptive fusion design.} The upper block tests learned routing; the lower block varies gate inputs and routing granularity.}
\label{tab:adaptive_fusion_design}
\centering
\scriptsize
\setlength{\tabcolsep}{2.2pt}
\renewcommand{\arraystretch}{1.08}
\begin{tabular*}{\columnwidth}{@{\extracolsep{\fill}}lcccc@{}}
\toprule
Setting & gIoU$\uparrow$ & $\mathrm{mIoU}_{\mathrm{all}}\uparrow$ & $\mathrm{mIoU}_{\mathrm{dyn}}\uparrow$ & $\mathrm{mIoU}_{\mathrm{sta}}\uparrow$ \\
\midrule
\multicolumn{5}{@{}l}{\textbf{\textit{Necessity of Adaptive Fusion}}} \\
Deterministic Average Merge & 86.34 & 58.29 & 27.72 & 81.22 \\
Deterministic Group-wise Merge & 86.98 & 58.75 & 28.04 & 81.78 \\
Adaptive Group-wise Fusion & 89.08 & 60.08 & 28.97 & 83.42 \\
\midrule
\multicolumn{5}{@{}l}{\textbf{\textit{Design of Adaptive Fusion}}} \\
Confidence-only (group-wise) & 86.72 & 58.56 & 27.81 & 81.63 \\
Probability-only (group-wise) & 87.26 & 58.95 & 28.18 & 82.03 \\
Prob.+Conf. (class-wise) & 88.18 & 59.44 & 28.55 & 82.61 \\
Prob.+Conf. (group-wise) & 89.08 & 60.08 & 28.97 & 83.42 \\
\bottomrule
\end{tabular*}
\end{table}

%% file: latex/tab/static-consistency.tex
\begin{table}[!tbp]
\caption{\textbf{Static consistency ablation.} We test whether a stable layout regularizer improves progressive reasoning. \rev{Batch-level consensus is the default consistency form.}}
\label{tab:static_consistency}
\centering
\scriptsize
\setlength{\tabcolsep}{2.8pt}
\renewcommand{\arraystretch}{1.08}
\begin{tabular*}{\columnwidth}{@{\extracolsep{\fill}}lcccc@{}}
\toprule
Setting & gIoU$\uparrow$ & $\mathrm{mIoU}_{\mathrm{all}}\uparrow$ & $\mathrm{mIoU}_{\mathrm{dyn}}\uparrow$ & $\mathrm{mIoU}_{\mathrm{sta}}\uparrow$ \\
\midrule
\multicolumn{5}{l}{\textbf{\textit{Necessity of Static Consistency}}} \\
No consistency & 86.10 & 58.34 & 28.20 & 80.95 \\
Static consistency (default) & \bestcell{89.08} & \bestcell{60.08} & \bestcell{28.97} & \bestcell{83.42} \\
\midrule
\multicolumn{5}{l}{\textbf{\textit{Design of Static Consistency}}} \\
\rev{Batch-level consensus} & \bestcell{89.08} & \bestcell{60.08} & \bestcell{28.97} & \bestcell{83.42} \\
\rev{Pairwise agreement} & 88.83 & 59.81 & 28.74 & 83.12 \\
\bottomrule
\end{tabular*}
\end{table}

%% file: latex/tab/loss-balance.tex
\begin{table}[t]
\caption{\textbf{\rev{Static-dynamic} loss balance.} We vary branch coefficients while keeping the model architecture fixed.}
\label{tab:loss_balance}
\centering
\scriptsize
\setlength{\tabcolsep}{2.6pt}
\renewcommand{\arraystretch}{1.08}
\begin{tabular*}{\columnwidth}{@{\extracolsep{\fill}}lcccccc@{}}
\toprule
Setting & $\lambda_{s}$ & $\lambda_{d}$ & gIoU$\uparrow$ & $\mathrm{mIoU}_{\mathrm{all}}\uparrow$ & $\mathrm{mIoU}_{\mathrm{dyn}}\uparrow$ & $\mathrm{mIoU}_{\mathrm{sta}}\uparrow$ \\
\midrule
Static-light  & 0.5 & 1.0 & 87.54 & 59.52 & 28.99 & 82.41 \\
Dynamic-light & 1.0 & 0.5 & 87.96 & 59.37 & 27.63 & 83.18 \\
Balanced      & 1.0 & 1.0 & \bestcell{89.08} & \bestcell{60.08} & 28.97 & \bestcell{83.42} \\
Dynamic-heavy & 1.0 & 2.0 & 87.83 & 59.82 & \bestcell{29.26} & 82.74 \\
Static-heavy  & 2.0 & 1.0 & 88.34 & 59.46 & 27.72 & 83.27 \\
\bottomrule
\end{tabular*}
\end{table}

%% file: latex/tab/loss-components.tex
\begin{table}[!tbp]
\caption{\textbf{Occupancy loss components.} CE, semantic-scaling, and Lovasz terms are varied for each supervision target.}
\label{tab:loss_terms}
\centering
\scriptsize
\setlength{\tabcolsep}{1.2pt}
\renewcommand{\arraystretch}{1.08}
\begin{tabular*}{\columnwidth}{@{\extracolsep{\fill}}lccccccc@{}}
\toprule
Setting & CE & Sem. & Lovasz & gIoU$\uparrow$ & $\mathrm{mIoU}_{\mathrm{all}}\uparrow$ & $\mathrm{mIoU}_{\mathrm{dyn}}\uparrow$ & $\mathrm{mIoU}_{\mathrm{sta}}\uparrow$ \\
\midrule
\multicolumn{8}{l}{\textbf{\textit{Static occupancy supervision}}} \\
CE only & \cmark &  &  & 86.80 & 58.48 & 26.89 & 82.18 \\
CE + Sem. & \cmark & \cmark &  & 87.89 & 59.22 & 27.08 & 83.33 \\
CE + Lovasz & \cmark &  & \cmark & 86.20 & 58.99 & 28.49 & 81.86 \\
CE + Sem. + Lovasz & \cmark & \cmark & \cmark & 89.08 & 60.08 & 28.97 & 83.42 \\
\midrule
\multicolumn{8}{l}{\textbf{\textit{Dynamic occupancy supervision}}} \\
CE only & \cmark &  &  & 86.78 & 57.33 & 25.49 & 81.21 \\
CE + Sem. & \cmark & \cmark &  & 87.54 & 58.33 & 27.34 & 81.57 \\
CE + Lovasz & \cmark &  & \cmark & 87.42 & 58.29 & 27.93 & 81.06 \\
CE + Sem. + Lovasz & \cmark & \cmark & \cmark & 89.08 & 60.08 & 28.97 & 83.42 \\
\midrule
\multicolumn{8}{l}{\textbf{\textit{Fusion-output occupancy supervision}}} \\
CE only & \cmark &  &  & 85.64 & 56.58 & 25.92 & 79.58 \\
CE + Sem. & \cmark & \cmark &  & 86.92 & 57.81 & 27.06 & 80.87 \\
CE + Lovasz & \cmark &  & \cmark & 86.56 & 57.56 & 27.47 & 80.13 \\
CE + Sem. + Lovasz & \cmark & \cmark & \cmark & 89.08 & 60.08 & 28.97 & 83.42 \\
\bottomrule
\end{tabular*}
\end{table}

%% file: latex/tab/accuracy-efficiency.tex
\begin{table}[!t]
\caption{\textbf{Accuracy--efficiency comparison.} \rev{Parameters are counted from our retrained configurations; latency is measured on a single NVIDIA RTX 4090 with batch size 1.}}
% TODO: confirm the latency measurement setup (GPU model and batch size).
\label{tab:capacity_efficiency}
\centering
\scriptsize
\setlength{\tabcolsep}{2.1pt}
\renewcommand{\arraystretch}{1.08}
\begin{tabular*}{\columnwidth}{@{\extracolsep{\fill}}lcccc@{}}
\toprule
Setting & Params. (M)$\downarrow$ & Lat. (ms)$\downarrow$ & $\mathrm{mIoU}_{\mathrm{all}}\uparrow$ & $\mathrm{mIoU}_{\mathrm{dyn}}\uparrow$ \\
\midrule
\multicolumn{5}{l}{\textbf{\textit{End-to-end efficiency}}} \\
\multicolumn{5}{l}{\textit{Multi-modal (C+L)}} \\
BEVFusion~\cite{BEVFusion} & 108.01 & 103.73 & 55.44 & 32.57 \\
\rev{ProSD-Occ (C+L)} & 87.02 & 164.42 & 65.87 & 37.92 \\
\midrule
\multicolumn{5}{l}{\textit{LiDAR-only (L)}} \\
VoxelNet~\cite{VoxelNet} & 73.25 & 75.43 & 53.87 & 31.13 \\
PointPillars~\cite{PointPillars} & 11.59 & 61.68 & 50.51 & 31.55 \\
\rev{ProSD-Occ (L)} & 30.24 & 109.55 & 63.66 & 33.87 \\
\midrule
\multicolumn{5}{l}{\textit{Camera-only (C)}} \\
BEVDet~\cite{BEVDet} & 100.66 & 41.92 & 45.14 & 21.19 \\
BEVFormer~\cite{BEVFormer} & 69.21 & 180.59 & 47.76 & 10.73 \\
BEVDepth~\cite{BEVDepth} & 100.66 & 57.43 & 46.81 & 19.64 \\
TPVFormer~\cite{TPVFormer} & 80.76 & 134.63 & 43.09 & 4.36 \\
SurroundOcc~\cite{SurroundOcc} & 80.04 & 44.40 & 50.77 & 11.37 \\
CONet~\cite{OpenOccupancy} & 100.74 & 53.23 & 49.25 & 18.93 \\
SparseOcc~\cite{SparseOcc} & 105.08 & 64.30 & 45.42 & 9.41 \\
GaussianFormer~\cite{GaussianFormer} & 36.32 & 65.50 & 43.63 & 10.45 \\
\rev{ProSD-Occ (C)} & 57.65 & 72.06 & 60.08 & 28.97 \\
\bottomrule
\end{tabular*}
\end{table}

%% file: latex/tab/distance-wise-dynamic.tex
\begin{table}[!tbp]
\caption{\textbf{Distance-wise dynamic occupancy of \nameplus{}.} \rev{Dynamic mIoU of the three \nameplus{} tracks is evaluated in BEV distance intervals from the infrastructure sensor origin.}}
\label{tab:distance_robustness}
\centering
\scriptsize
\setlength{\tabcolsep}{3.6pt}
\renewcommand{\arraystretch}{1.08}
\begin{tabular*}{\columnwidth}{@{\extracolsep{\fill}}lccc@{}}
\toprule
Setting & \multicolumn{3}{c}{$\mathrm{mIoU}_{\mathrm{dyn}}\uparrow$} \\
\cmidrule(lr){2-4}
& 0--20\,m & 20--40\,m & 40--60\,m \\
\midrule
ProSD-Occ (C) & 29.66 & 31.90 & 8.24 \\
ProSD-Occ (L) & 35.36 & 34.08 & 21.22 \\
ProSD-Occ (C+L) & 37.58 & 39.60 & 23.28 \\
\bottomrule
\end{tabular*}
\end{table}

%% file: latex/tab/background-overfitting-robustness.tex
\begin{table}[!tbp]
\caption{\textbf{Robustness to static-background overfitting.} \rev{We
re-anchor the evaluation frame} with controlled translations or yaw rotations while keeping the
physical scene unchanged, so coordinate-specific static-background shortcuts
become unreliable. Each entry reports $\mathrm{mIoU}_{\mathrm{dyn}}$.}
\label{tab:background_overfitting}
\centering
\scriptsize
\setlength{\tabcolsep}{2.2pt}
\renewcommand{\arraystretch}{1.08}
\begin{tabularx}{\columnwidth}{@{}>{\raggedright\arraybackslash}X C{1.08cm}C{1.08cm}C{1.08cm}C{1.08cm}@{}}
\toprule
Setting & \multicolumn{4}{c}{$\mathrm{mIoU}_{\mathrm{dyn}}\uparrow$} \\
\cmidrule(lr){2-5}
& Clean & 0.4\,m & 0.8\,m & 1.2\,m \\
\midrule
\multicolumn{5}{l}{\textbf{\textit{Translation re-anchoring}}} \\
\rev{Plain (C)} & 23.48 & 23.35 & 22.36 & 21.64 \\
ProSD-Occ (C) & 28.97 & 28.32 & 27.80 & 26.95 \\
ProSD-Occ (L) & 33.87 & 31.05 & 27.70 & 29.34 \\
ProSD-Occ (C+L) & 37.92 & 36.42 & 33.51 & 33.91 \\
\midrule
Setting & \multicolumn{4}{c}{$\mathrm{mIoU}_{\mathrm{dyn}}\uparrow$} \\
\cmidrule(lr){2-5}
& Clean & $0.4^\circ$ & $0.8^\circ$ & $1.2^\circ$ \\
\midrule
\multicolumn{5}{l}{\textbf{\textit{Rotation re-anchoring}}} \\
\rev{Plain (C)} & 23.48 & 22.79 & 21.93 & 21.60 \\
ProSD-Occ (C) & 28.97 & 27.77 & 26.55 & 26.08 \\
ProSD-Occ (L) & 33.87 & 32.43 & 32.27 & 25.92 \\
ProSD-Occ (C+L) & 37.92 & 36.20 & 34.40 & 28.85 \\
\bottomrule
\end{tabularx}
\end{table}

%% file: latex/fig/robustness-curves.tex
\begin{figure}[!t]
  \centering
  \includegraphics[width=\columnwidth]{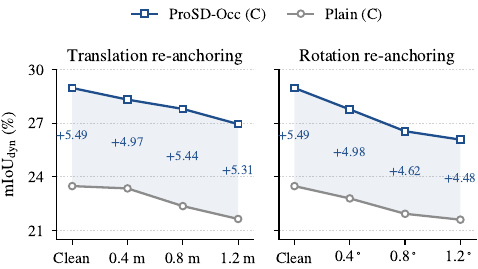}
  \caption{\textbf{Robustness comparison between \nameplus{} (C) and Plain (C).}
  \rev{The two panels report absolute dynamic mIoU under translation and
  rotation re-anchoring; the shaded band and the annotated values denote the
  margin of \nameplus{} (C) over Plain (C) at each perturbation level.}}
  \label{fig:robustness_c_prosd_vs_plain}
\end{figure}

%% file: latex/5-conclusion.tex
This paper studies infrastructure-side semantic occupancy, where fixed
roadside sensors repeatedly observe stable layouts while dynamic participants
remain sparse, local, and transient. We introduce InfraOcc, a real-world
benchmark with dense voxel annotations\rev{, unified camera-only, LiDAR-only,
and multi-modal protocols, and statistics that quantify the structural
static-dynamic asymmetry of fixed-view scenes}, and propose \nameplus{}, a progressive static-to-dynamic
framework that first models persistent layout and then exposes residual dynamic
evidence under static-confidence guidance. Experiments on InfraOcc show
consistent improvements in overall and dynamic occupancy across sensing
regimes, and diagnostic analyses indicate that the gain comes from organizing
fixed-view static-dynamic evidence rather than from sensor-specific feature
construction alone. These results suggest that roadside occupancy should be
evaluated and modeled around its fixed-view scene structure. \rev{Meanwhile,
long-range dynamic occupancy remains bounded by the available geometric
evidence, and our study is confined to keyframe-level prediction. Future work
will extend InfraOcc toward occupancy flow and exploit the fixed viewpoint
for temporal fusion, where the shared roadside coordinate system naturally
supports long-term scene memory.}